%% file: main.tex
\pdfoutput=1
\ifdefined\pdfsuppressptexinfo\pdfsuppressptexinfo=-1\fi
\documentclass{article}

\usepackage{iclr2027_conference,times}

\input{math_commands.tex}

\usepackage{amsmath}
\usepackage{amssymb}
\usepackage{pifont}
\usepackage{marvosym}
\usepackage{booktabs}
\usepackage{colortbl}
\usepackage{graphicx}
\usepackage{wrapfig}
\usepackage{capt-of}
\usepackage{needspace}
\usepackage{float}
\usepackage{placeins}
\usepackage{afterpage}
\usepackage{hyperref}
\hypersetup{hidelinks}
\usepackage{url}

\definecolor{ReMAPBaseline}{HTML}{B7D3E1}
\definecolor{ReMAPLocal}{HTML}{BFE0B5}
\definecolor{ReMAPSecondary}{HTML}{F6D2A0}
\definecolor{ReMAPGlobal}{HTML}{ABA6E0}
\definecolor{ReMAPSignal}{HTML}{8DE3FE}
\definecolor{ReMAPReplay}{HTML}{B8E4E7}
\definecolor{ReMAPDegraded}{HTML}{FDD0BD}
\definecolor{ReMAPPrimary}{HTML}{FCBFC0}
\definecolor{ReMAPFirst}{HTML}{A8C8F0}
\definecolor{ReMAPSecond}{HTML}{DCEAFB}

\makeatletter
\newcommand{\finishwrap}{%
  \par
  \ifnum\c@WF@wrappedlines>\@ne
    \vskip\dimexpr\c@WF@wrappedlines\baselineskip-\baselineskip\relax
  \fi
  \WFclear}
\newcommand{\continuedfigurecaption}[2]{%
  \@makecaption{Figure~\ref{#1} (continued)}{#2}}
\makeatother

\definecolor{ReMAPTitleBlue}{HTML}{326A9D}
\definecolor{ReMAPCiteBlue}{HTML}{4381BA}
\definecolor{GoogleBlue}{HTML}{4285F4}
\definecolor{GoogleRed}{HTML}{EA4335}
\definecolor{GoogleYellow}{HTML}{FBBC05}
\definecolor{GoogleGreen}{HTML}{34A853}
\newcommand{\ReMAPcolor}{{\color{GoogleBlue}R}{\color{GoogleRed}e}{\color{GoogleYellow}M}%
  {\color{GoogleBlue}A}{\color{GoogleGreen}P}}
\newcommand{\authorname}[2]{\mbox{#1\textsuperscript{\mdseries #2}}}
\newcommand{\affilsep}{\hspace{0.8em}}
\providecommand{\showlogos}{1}
\newcommand{\affilentry}[3]{\mbox{\textsuperscript{#1}\ifnum\showlogos=1
  \,\raisebox{-0.3em}{\includegraphics[height=1.25em]{figures/logos/#2}}\,\fi #3}}
\usepackage{tcolorbox}
\definecolor{ReMAPAbstractGray}{HTML}{F2F3F5}
\newtcolorbox{abstractpanel}{colback=ReMAPAbstractGray, colframe=ReMAPAbstractGray, boxrule=0pt,
  arc=2pt, boxsep=0pt, left=9pt, right=9pt, top=7pt, bottom=7pt}
\newcommand{\email}[1]{\href{mailto:#1}{{\fontfamily{lmtt}\selectfont #1}}}

\makeatletter
\newcommand{\titlenote}[1]{\begingroup
  \renewcommand{\@makefntext}[1]{\noindent##1}%
  \renewcommand{\thefootnote}{}%
  \footnotetext{\raggedright #1}\endgroup}
\def\section{\@startsection {section}{1}{\z@}{-2.0ex plus
    -0.5ex minus -.2ex}{1.5ex plus 0.3ex
minus0.2ex}{\large\sc\raggedright\color{ReMAPTitleBlue}}}
\def\subsection{\@startsection{subsection}{2}{\z@}{-1.8ex plus
-0.5ex minus -.2ex}{0.8ex plus .2ex}{\normalsize\sc\raggedright\color{ReMAPTitleBlue}}}
\def\subsubsection{\@startsection{subsubsection}{3}{\z@}{-1.5ex
plus      -0.5ex minus -.2ex}{0.5ex plus
.2ex}{\normalsize\sc\raggedright\color{ReMAPTitleBlue}}}
\makeatother

\iclrfinalcopy

\hypersetup{
  colorlinks=true, citecolor=ReMAPCiteBlue, linkcolor=ReMAPTitleBlue, urlcolor=., filecolor=.,
  pdftitle={ReMAP: Restoring the Perceptual Cycle with Reasoning-Time Latent Visual Memory},
  pdfauthor={Hao Jiang, Zhanyu Guo, Chenwei Wu, Yichen Guo, Qizhe Zhang, Junchi Yao, Jixian Wu, Jinhao You, Kai Tang, Jiajun Cao, Tinghao Wang, Mengyu Wang, Leo Anthony Celi, Shanghang Zhang}}

\begin{document}

\raggedbottom

\thispagestyle{empty}
\lhead{Preprint. Under review.}

\begin{center}
  {\LARGE\bfseries\ReMAPcolor: Restoring the Perceptual Cycle\\
    with Reasoning-Time Latent Visual Memory\par}
  \vspace{16pt}
  {\bfseries
    \authorname{Hao Jiang}{1*}, \authorname{Zhanyu Guo}{1*},
    \authorname{Chenwei Wu}{2\dag}, \authorname{Yichen Guo}{3,4\dag},
    \authorname{Qizhe Zhang}{4},\\
    \authorname{Junchi Yao}{5},
    \authorname{Jixian Wu}{6}, \authorname{Jinhao You}{1},
    \authorname{Kai Tang}{4}, \authorname{Jiajun Cao}{4},\\
    \authorname{Tinghao Wang}{1}, \authorname{Mengyu Wang}{7},
    \authorname{Leo Anthony Celi}{8}, \authorname{Shanghang Zhang}{4\Letter}\par}
  \vspace{5pt}
  {\small
    \affilentry{1}{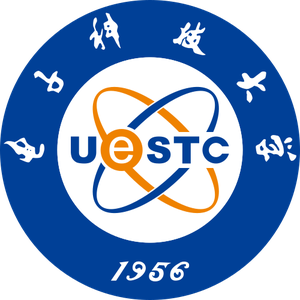}{University of Electronic Science and Technology of China},\affilsep
    \affilentry{2}{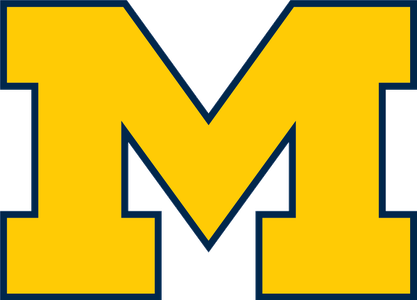}{University of Michigan, Ann Arbor},\affilsep
    \affilentry{3}{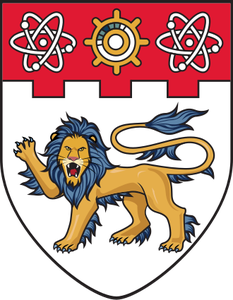}{Nanyang Technological University},\affilsep
    \affilentry{4}{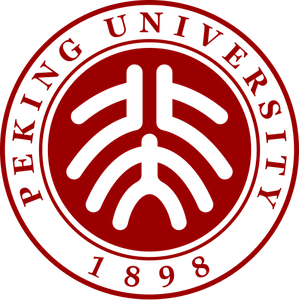}{Peking University},\affilsep
    \affilentry{5}{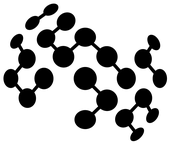}{Mohamed bin Zayed University of Artificial Intelligence},\affilsep
    \affilentry{6}{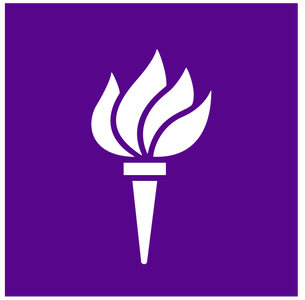}{New York University},\affilsep
    \affilentry{7}{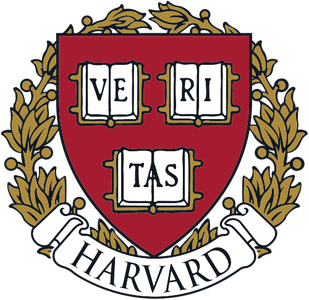}{Harvard University},\affilsep
    \affilentry{8}{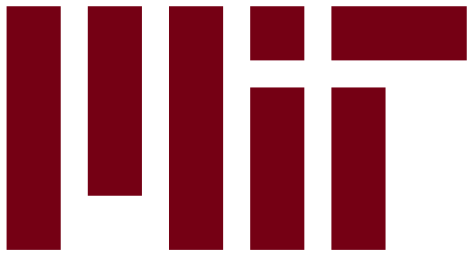}{Massachusetts Institute of Technology}\par}
\end{center}
\vspace{4pt}
\titlenote{\textsuperscript{*}Equal contribution.\quad
  \textsuperscript{\dag}Project leader.\quad
  \textsuperscript{\Letter}Corresponding author.\\
  Correspondence to: \mbox{Hao Jiang (\email{2024010906013@std.uestc.edu.cn})},
  \mbox{Shanghang Zhang (\email{shanghang@pku.edu.cn})}.}

\renewenvironment{abstract}{\begin{abstractpanel}%
  \noindent{\color{ReMAPTitleBlue}\textbf{Abstract:}}\ \ignorespaces}{\end{abstractpanel}}
\input{sections/00_abstract}

\input{figures/figure1/figure1}

\input{sections/01_introduction}
\input{sections/02_related_work}
\input{sections/03_method}

\input{sections/04_experiments}
\input{sections/05_limitations}
\input{sections/05_conclusion}

\subsection*{AI use statement}
\input{statements/ai_use}

\subsection*{Ethics statement}
\input{statements/ethics}

\subsection*{Reproducibility statement}
\input{statements/reproducibility}

\FloatBarrier
\bibliography{references}
\bibliographystyle{iclr2027_conference}

\appendix
\input{sections/90_appendix}

\end{document}

%% file: math_commands.tex
\usepackage{amsmath,amsfonts,bm}

\def\eqref#1{equation~\ref{#1}}

\def\1{\bm{1}}

\DeclareMathAlphabet{\mathsfit}{\encodingdefault}{\sfdefault}{m}{sl}
\SetMathAlphabet{\mathsfit}{bold}{\encodingdefault}{\sfdefault}{bx}{n}



%% file: sections/00_abstract.tex
\begin{abstract}
As multimodal large language models (MLLMs) reason for longer, attention to the
initial visual input diminishes, weakening visual grounding. Visual memory
reintroduces visual evidence during reasoning. We
conduct a controlled analysis of visual memory along three axes: curation,
organization, and access. We find that local evidence
benefits from global context, compact latent representations balance accuracy
and visual-context cost, and the utility of memory access depends on the
reasoning state.
Guided by these findings, we propose \textbf{\ReMAPcolor}
(\textbf{\underline{\textcolor{GoogleBlue}{R}\textcolor{GoogleRed}{e}}asoning-Time
\underline{\textcolor{GoogleYellow}{M}}emory-\underline{\textcolor{GoogleBlue}{A}}ugmented
\underline{\textcolor{GoogleGreen}{P}}erception}), which couples two
complementary latent memories:
a static, question-conditioned Global memory that preserves scene and
cross-image context, and a dynamic Local memory that uses this
context as an anchor while selecting and re-encoding region-level evidence
according to the current reasoning state and the question.
Both memories return compact latent tokens that are inserted into the reasoning
sequence, and a reinforcement-learning access policy trained with branched
rollouts decides when to continue reasoning or invoke Global or Local memory.
On ten benchmark families, ReMAP outperforms prior visual-memory methods on all
four multi-image benchmarks, exceeding the strongest prior results on MuirBench
and MIMIC by 8.38 and 14.84 percentage points, respectively. Across four
backbone families, enabling memory access improves over the same trained model
with memory disabled, and on shared V*Bench, CV-Bench-2D, and MuirBench
questions ReMAP reduces the visual tokens entering the reasoning sequence by
51.0--76.8\% relative to the native-resolution backbone. Further analyses show
that Global and Local memory form distinct yet complementary latent
representations and play coordinated roles in interleaved vision-language
reasoning. Together, these components restore the perceptual cycle by letting
the reasoning state trigger targeted visual retrieval, with the retrieved
evidence guiding subsequent reasoning. Code coming soon.
\end{abstract}

%% file: figures/figure1/figure1.tex
\begin{figure}[!t]
  \centering
  \includegraphics[width=0.8\linewidth]{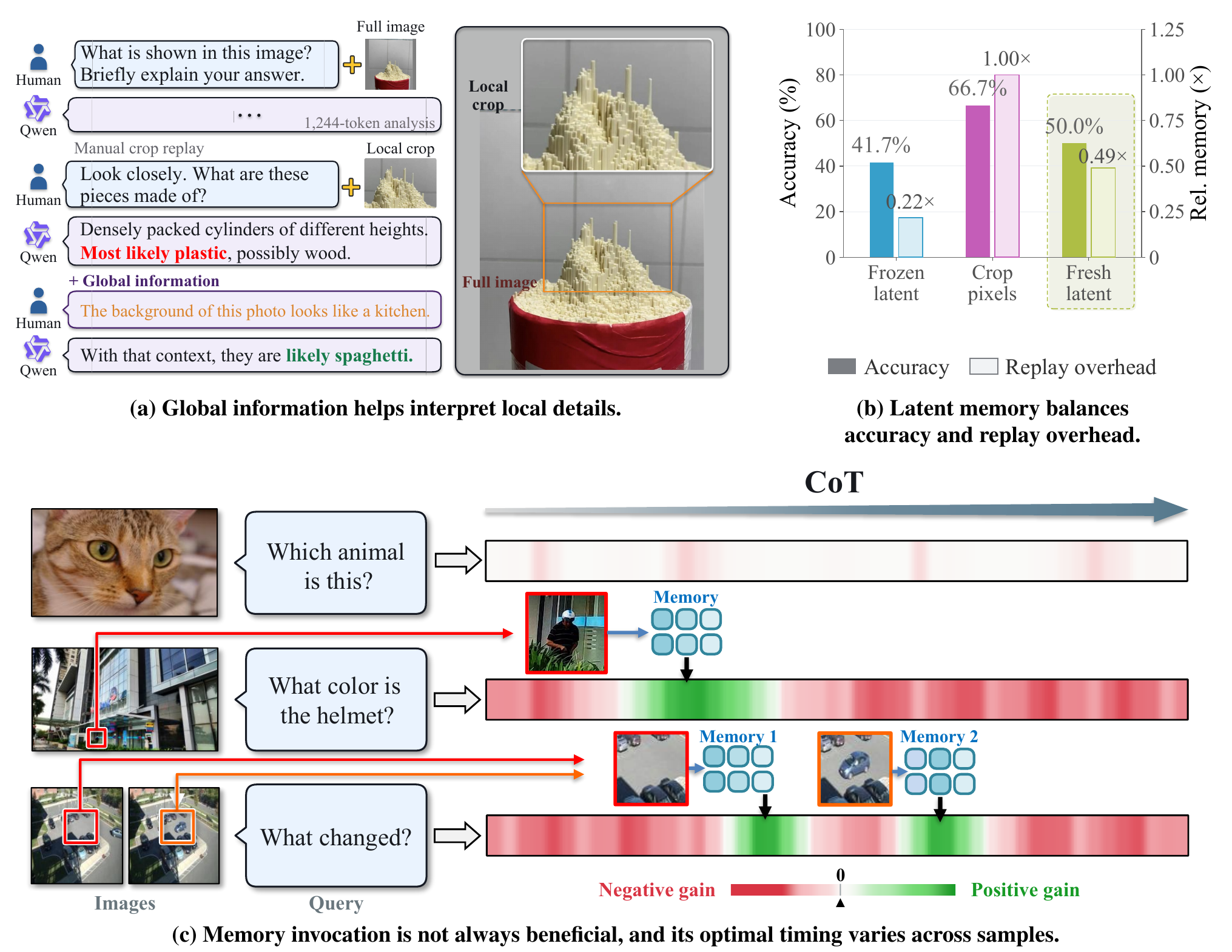}
  \caption{\textbf{Investigating memory curation, organization, and access.} \textbf{(a)} An interaction with Qwen2.5-VL-7B-Instruct on a MuirBench example illustrates how global context helps resolve ambiguous local evidence. \textbf{(b)} A comparison of frozen latents, crop pixels, and fresh latents on a random 25\% subset of V*Bench examines the trade-off between accuracy and incremental GPU-memory overhead. \textbf{(c)} An easy recognition example and cases from V*Bench and MuirBench show how the need for memory and its invocation timing vary across queries and reasoning states.}
  \label{fig:quantified-bottlenecks}
\end{figure}

%% file: sections/01_introduction.tex
\section{Introduction}
\label{sec:introduction}

Vision--language models are increasingly applied to visual tasks
covering larger visual contexts and longer reasoning trajectories. These tasks
require models to draw on visual evidence at different reasoning stages such as
inspecting fine-grained details, comparing observations across images, or
verifying intermediate conclusions. As visual inputs and generated text
accumulate, maintaining access to relevant visual evidence is
increasingly challenging. Recent studies report visual forgetting, where
attention to visual inputs diminishes during extended
generation~\citep{sun2025takealong,liu2025morethinking}, highlighting the
difficulty of sustaining visual grounding throughout reasoning.

Visual memory methods have emerged as a natural solution to retain visual
evidence during reasoning. Existing approaches explore several routes with
distinct trade-offs: reinspecting source pixels offers fine-grained details but
requires extra image processing~\citep{zheng2026deepeyes,su2025pixelreasoner,jiang2025vlmr3}, replaying
encoded features reuses existing representations but limits available evidence
to their initial coverage~\citep{chu2025lookagain,gao2025interleaved,zhang2026cmmcot},
constructing latent memories provides compact representations at the cost of
interpretability~\citep{zhang2026memgen,chen2026flexmem,yu2026vismem}. These approaches differ in evidence
selection, representation, and delivery timing. We therefore ask: \textbf{what
design choices of a visual memory system truly improve visual reasoning?}

We characterize visual memory along three design axes:
\textbf{curation} (what evidence enters memory),
\textbf{organization} (how evidence is represented), and
\textbf{access} (when to access memory and which memory to use).
Under this framework, we conduct preliminary experiments combining
qualitative analysis with controlled empirical evaluations, as illustrated in
Figure~\ref{fig:quantified-bottlenecks}.
These analyses yield three key findings:
\ding{182} Memory \textbf{curated} from local visual evidence alone can be
insufficient for contextual reasoning, as illustrated in
Figure~\ref{fig:quantified-bottlenecks}(a).
\ding{183} Memory \textbf{organization} affects both reasoning accuracy and
computational overhead, as illustrated by the trade-off between image-crop replay
and compact latent representations in Figure~\ref{fig:quantified-bottlenecks}(b).
\ding{184} Reintroducing visual evidence is not
always beneficial, and its utility depends on
when it is \textbf{access}ed during reasoning
(Figure~\ref{fig:quantified-bottlenecks}(c)).

Guided by these findings, we propose \textbf{ReMAP} (\emph{Reasoning-Time
Memory-Augmented Perception}). For \textbf{memory curation}, ReMAP combines Global
memory, which preserves scene and cross-image context from the input images and
question, with Local memory, which selects and re-encodes image regions according
to the current reasoning state. For \textbf{memory organization}, Global and
Local memory return compact latent tokens that are inserted into the reasoning
sequence to balance reasoning accuracy and computational overhead.
For \textbf{memory access}, we treat memory invocation as an agentic
tool call so that the model can obtain relevant visual evidence when 
needed. We leverage branched rollouts before each invocation to learn whether,
when, and which memory to invoke. These rollouts compare the outcomes of
requesting memory versus continuing from the same reasoning state and teach
when memory helps. Inference follows a single trajectory.

We evaluate ReMAP against existing visual memory methods on ten benchmark
families spanning single-image perception, multi-image reasoning, and broader
visual understanding. ReMAP achieves state-of-the-art overall performance
across the benchmark suite. Notably, ReMAP significantly improves multi-image
reasoning, outperforming existing visual-memory methods on all four benchmarks
and exceeding the strongest prior results on MuirBench and MIMIC by 8.38 and
14.84 percentage points, respectively. Beyond these aggregate gains, our
analyses characterize \mbox{reasoning-time} visual memory: ReMAP's memory access improves
performance across four backbone families, Global and Local evidence provide
complementary benefits, and compact latent returns reduce visual-context cost
while maintaining leading accuracy. Further analyses make Global and Local
memory interpretable through their distinct yet complementary latent
representations and coordinated roles in interleaved vision-language
reasoning. Our contributions are threefold:

\begin{itemize}
  \item[\ding{182}] \textbf{We identify three principles for reasoning-time visual memory:}
  contextualize local evidence with global context, compress representations,
  and adapt access to the evolving reasoning state.

  \item[\ding{183}] \textbf{We introduce ReMAP, a framework for reasoning-time visual memory.}
  ReMAP combines query-conditioned Global context with state-guided Local
  evidence in compact latent returns and learns when and which memory to invoke
  through branched rollouts.

  \item[\ding{184}] \textbf{We demonstrate strong multi-image reasoning with interpretable memory roles.}
  ReMAP outperforms prior visual-memory methods on all four multi-image
  benchmarks, while representation and trajectory analyses reveal the
  complementary functions of Global and Local memory.
\end{itemize}

%% file: sections/02_related_work.tex
\section{Related Work}
\label{sec:related-work}

Memory curation spans pixel reacquisition, feature replay, and learned memory
formation, with distinct trade-offs between evidence coverage and processing
overhead \citep{zheng2026deepeyes,su2025pixelreasoner,zhang2026memgen}. Memory
organization represents evidence through images, text, encoded features, or
latent tokens, with different trade-offs in interpretability and context cost
\citep{hu2024visualsketchpad,zhang2026lookondemand,huang2026persistent,yu2026vismem}.
Memory access uses heuristic triggers, uncertainty criteria, or learned policies
\citep{sun2025takealong,liu2026dapicot,peng2026reground}, and related policy
optimization methods improve tool use and visual reasoning through training
feedback \citep{dong2025arpo,li2026rethinking}. ReMAP bridges these dimensions by
combining pixel reacquisition with compact latent memory and preserving scene
and cross-image context to interpret newly acquired local evidence. Its access
training branches before memory invocation to compare requesting memory with
continuing from the same reasoning state, providing feedback on the utility of
memory access. A detailed review and comparison are provided in
Appendix~\ref{app:related-work}.

%% file: sections/03_method.tex
\section{Reasoning-Time Memory-Augmented Perception}
\label{sec:method}

In this section, we introduce ReMAP, which interleaves autoregressive reasoning
with access to latent visual memory to restore the perceptual cycle.
Figure~\ref{fig:remap-overview} provides an overview of the framework. Given
images $\mathcal I=\{I_i\}_{i=1}^{N}$ and a question $Q$, the model begins
reasoning from a coarse visual prefix $V^c=E_c(\mathcal I)$. During reasoning,
the model can continue generating text or invoke Global or Local latent
memory. After each memory return, the model resumes reasoning with the
retrieved evidence. Specifically, Section~\ref{sec:method-curation} describes
how Global and Local memories are formed and integrated into the reasoning
sequence.
Section~\ref{sec:method-learning} describes memory access and its training policy.

\input{figures/method-overview/figure_method_overview}

\subsection{Memory Curation and Organization}
\label{sec:method-curation}

\paragraph{Global memory.}
\label{sec:method-global}
Global memory provides a compact representation of image content and, for
multi-image inputs, relationships across images. We construct it in two stages:
the Image Weaver summarizes each image, and the Relation Weaver integrates
these summaries into relational context. Both stages are conditioned on the
input question.

Let $V_i$ denote the encoded visual features of image $I_i$. We use two sets of
learnable query tokens, $Q_G$ and $Q_R$, to aggregate per-image content and
relational context, respectively. The corresponding modules,
$W_{\mathrm{img}}$ and $W_{\mathrm{rel}}$, apply cross-attention followed by
feed-forward layers. The Image Weaver attends to each image's features
separately, while the Relation Weaver attends to the concatenated image
summaries. Both include an encoded representation of the question in their
attention context, denoted by the conditioning on $Q$ below:
\begin{equation}
  Z_i^{\mathrm{img}}=W_{\mathrm{img}}(Q_G,V_i;Q),
  \qquad
  \mathcal Z^{\mathrm{rel}}
  =W_{\mathrm{rel}}\bigl(Q_R,[Z_1^{\mathrm{img}};\cdots;Z_N^{\mathrm{img}}];Q\bigr).
  \label{eq:image-weaver}
\end{equation}
The Global memory block
$Z_G=[Z_1^{\mathrm{img}};\cdots;Z_N^{\mathrm{img}};\mathcal Z^{\mathrm{rel}}]$
concatenates the image summaries and relational context.
Constructed and cached before decoding, $Z_G$ depends on the input images and
question and remains fixed throughout reasoning. When the decoder generates
\texttt{<mem\_g>}, the runtime inserts $Z_G$ followed by
\texttt{<mem\_g\_end>} and resumes reasoning.

\paragraph{Local memory.}
\label{sec:method-local}
Figure~\ref{fig:local-memory-details}(a) shows how Local memory dynamically
retrieves fine-grained evidence as reasoning evolves.
At state $h_t$, the Source Selector uses the layer-12 request state
$h_t^{12}$ to select images $\mathcal S_t$, and Twig uses the layer-18 prefix
states $H_{<t}^{18}$ to localize a crop in each selected image. Twig extends
the ROI predictor of SD-RPN~\citep{shi2025sdrpn} and
Q-Zoom~\citep{shi2026qzoom} from the input question to the current reasoning
prefix, so successive requests can localize different regions. A frozen ROI
encoder maps the crops to region features $R_{t,i}$. The Local Weaver takes
the cached Global memory $Z_G$ as an additional input while compressing each
region:
\begin{equation}
  \widetilde Z_{t,i}^{L}
  =W_L(h_t,R_{t,i};Z_G),
  \qquad
  Z_L=\big[\widetilde Z_{t,i}^{L}\big]_{i\in\mathcal S_t}.
  \label{eq:local-memory}
\end{equation}
$Z_G$ anchors each crop in the image-level and cross-image context that local
evidence alone can miss, consistent with our first finding.
A \texttt{<mem\_l>} request similarly inserts $Z_L$ followed by
\texttt{<mem\_l\_end>} before generation resumes. The concatenated block
contains $K$ Local latents.

\input{figures/local-detail/figure_local_detail}

\subsection{Memory Access}
\label{sec:method-learning}

Latent memory keeps retrieved visual evidence within a fixed token budget.
Replaying whole images at every request would repeatedly expand the context
and compete with the ongoing reasoning sequence. Moreover, memory does not
help at every step, since its value depends on when it is accessed, consistent
with our third finding. The content of a return can be supervised directly,
whereas the value of its timing appears only in later reasoning. We therefore
separate memory formation from memory access. Stage I uses supervised fine-tuning to form
useful memory and warm up request handling, while Stage II uses reinforcement
learning to learn whether and when to invoke Global or Local memory.

\paragraph{Stage I: Supervised learning and warm-up.}
Stage I isolates memory formation from access-policy learning, so that each
return localizes the right region and carries evidence that supports
subsequent reasoning. The vision encoder, pretrained LLM, and decoder LoRA
remain frozen throughout this stage. Existing datasets lack ROI annotations
that change along the reasoning chain, so we annotate reasoning steps with
target regions, together with reference CoTs from a teacher VLM, per-image
facts, cross-image relations, and answers. We first optimize memory quality by
training the Image and Relation Weavers and then the Local Weaver. Their
returns are inserted into teacher CoTs, and the continuation cross-entropy
$\mathcal L_{\mathrm{suffix}}$ through the frozen LLM, together with fact,
relation, and answer supervision, trains them to encode evidence for the
teacher's subsequent reasoning. We then train Twig to localize the annotated
regions with $\mathcal L_{\mathrm{ROI}}$. Because student states differ from
teacher states, we next train continuation on student rollouts, where the
teacher reviews the student's prefixes and crops to provide continuation and
ROI targets for the same two losses. Finally, with the formation modules
fixed, a warm-up lets the model emit the memory special tokens, updating only
the input embeddings of the four markers and the output-layer rows that score
\texttt{<mem\_g>} and \texttt{<mem\_l>}. Figure~\ref{fig:local-memory-details}(b) illustrates these
signals, and Appendix~\ref{app:formation-objectives} gives the objectives.

\finishwrap

\input{tables/main_results}

\paragraph{Stage II: Reinforcement learning for memory access.}
Guided by our third finding, Stage II makes memory access helpful by learning
when and which memory to request, with formation modules frozen. At a request
boundary, the decoder emits a Global or Local request token or continues
generating text. A trajectory's final reward cannot show whether a given request helped.
We therefore branch from a shared prefix and compare
requesting with continuing at the same state. Following
ARPO~\citep{dong2025arpo}, each root branches at most
once, at the state maximizing
\begin{equation}
  u(s_t,a_t)=1-p_t(a_t)^m,
  \label{eq:crd}
\end{equation}
where $p_t(a_t)$ is the probability of the realized route $a_t$, with
$p_t(\mathrm N)=1-p_t(\mathrm G)-p_t(\mathrm L)$ for no request, and $m$ is
the number of sibling rollouts. This score is the probability that at least
one sibling takes a different route, so the branch goes where siblings likely
diverge from the root, as illustrated in
Figure~\ref{fig:local-memory-details}(c). A sibling $i$ that requests memory at
branch $b$ receives the invocation gain $\Delta_i^{\mathrm{call}}=R_i-B_b$, where
$R_i$ is its final-answer reward and $B_b$ is the mean reward of siblings that
continue without a request. These siblings may still request
memory later, so the gain measures the value of requesting now. Normalized
invocation gains update the request-token embeddings and output rows, and
group-relative final-answer advantages~\citep{shao2024deepseekmath} update the
decoder LoRA. Appendix~\ref{app:access-objectives} gives the objectives.

%% file: figures/method-overview/figure_method_overview.tex
\begin{figure}[t]
  \centering
  \includegraphics[width=0.95\linewidth]{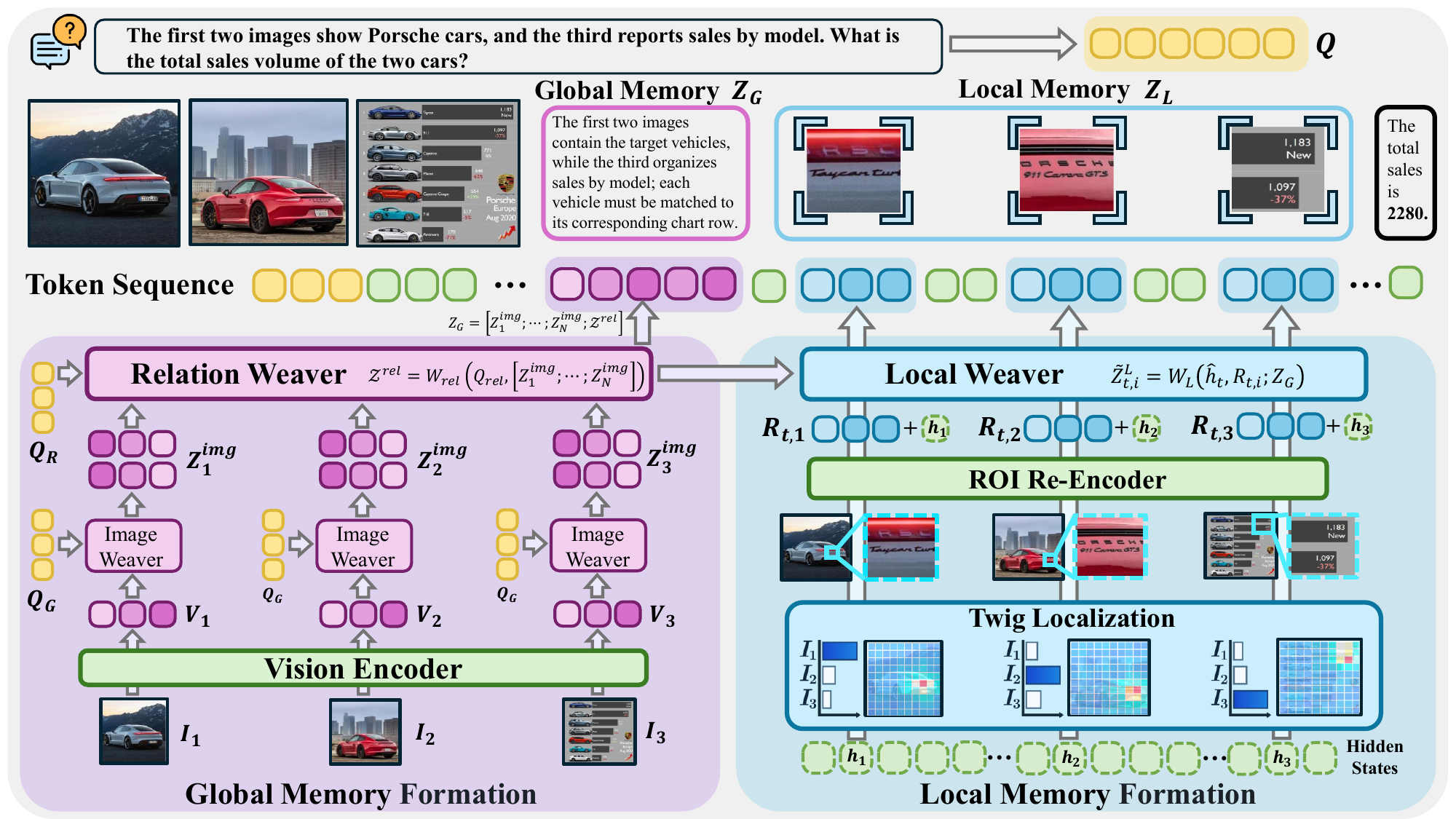}
  \caption{\textbf{Overview of ReMAP.}
  Global memory forms a question-conditioned summary of individual images and
  their relations before decoding. During reasoning, a Local request uses the
  current state to select sources and regions, re-encodes selected pixels, and
  compresses them with cached Global context. Global and Local returns are
  inserted as continuous latent blocks after their request markers, followed
  by end markers before autoregressive decoding resumes.}
  \label{fig:remap-overview}
\end{figure}

%% file: figures/local-detail/figure_local_detail.tex
\begin{figure}[t]
  \centering
  \includegraphics[width=0.95\linewidth,clip]{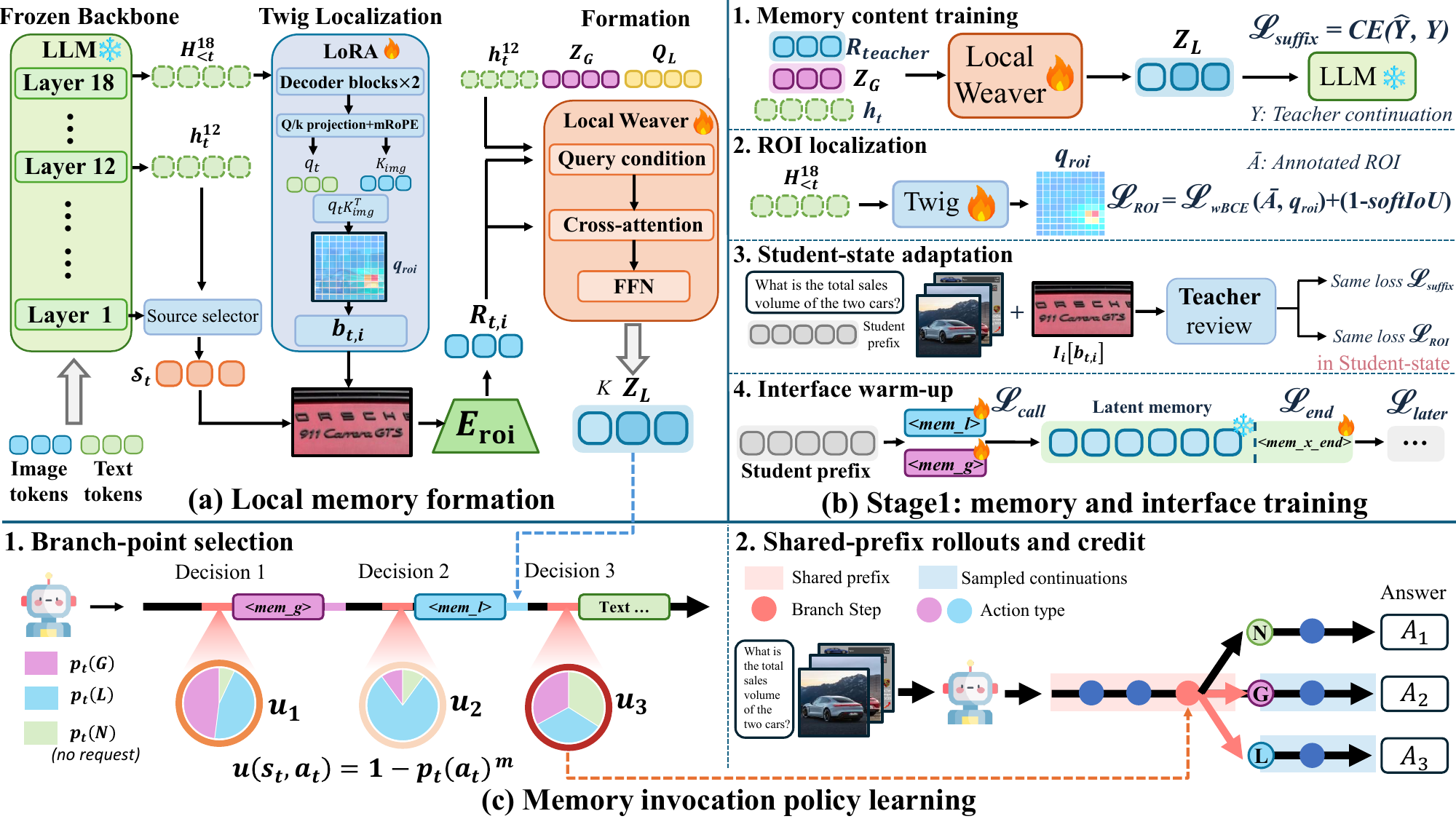}
  \caption{\textbf{Local memory and adaptive invocation.}
  (a) The request state $h_t^{12}$ selects source images $\mathcal S_t$, Twig
  uses the prefix states $H_{<t}^{18}$ to localize a crop in each, and the
  Local Weaver compresses the crops with the Global anchor $Z_G$ into $K$
  Local latents.
  (b) Stage I trains the Local Weaver with $\mathcal L_{\mathrm{suffix}}$ and
  Twig with $\mathcal L_{\mathrm{ROI}}$, adapts both to teacher-reviewed
  student states, and warms up the memory tokens.
  (c) Each root branches where siblings are most likely to take a different
  route, and the invocation gain compares requesting memory with the default
  continuation $\mathrm N$ at the same state.}
  \label{fig:local-memory-details}
\end{figure}

%% file: tables/main_results.tex
\begin{table}[!t]
\centering
\scriptsize
\setlength{\tabcolsep}{2.0pt}
\setlength{\abovecaptionskip}{0pt}\setlength{\belowcaptionskip}{4pt}
\caption{\textbf{Comparison with existing visual memory methods across ten benchmark families.} Results cover single-image perception, multi-image reasoning, and broader visual understanding. MIMIC Avg. is the mean across four subsets. The top-2 results in each column are highlighted as \protect\colorbox{ReMAPFirst}{first} and \protect\colorbox{ReMAPSecond}{second}, with the best results also shown in \textbf{bold}.}
\label{tab:main-results}
\resizebox{\textwidth}{!}{%
\renewcommand{\arraystretch}{1.2}
\begin{tabular}{lcccccccccccc}
\toprule
& \multicolumn{5}{c}{Single-image evidence acquisition}
& \multicolumn{4}{c}{Multi-image relational reasoning}
& \multicolumn{3}{c}{Held-out generalization} \\
\cmidrule(lr){2-6}\cmidrule(lr){7-10}\cmidrule(lr){11-13}
Method & V$^\star$ & CV-2D & CV-3D & HR-4K & HR-8K
& BLINK & Muir & MMSI & \shortstack{MIMIC\\Avg.} & RWQA & MathVista & MMVet \\
Native & 76.44 & 74.41 & 76.33 & 72.88 & 67.75 & 56.62 & 57.85 & 28.10 & 45.17 & 67.97 & 59.90 & 46.06 \\
Coarse & 67.54 & 73.10 & 75.42 & 61.13 & 54.62 & 57.79 & \cellcolor{ReMAPSecond}58.15 & 28.40 & 43.56 & 67.19 & 59.50 & 47.06 \\
\midrule
\multicolumn{13}{l}{\textit{Pixel reinspection}} \\
Q-Zoom & 79.06 & 76.35 & 74.08 & 73.00 & 66.63 & 53.97 & 50.38 & 25.40 & 42.29 & 70.85 & 54.60 & 47.43 \\
DeepEyes & \cellcolor{ReMAPFirst}\textbf{84.29} & \cellcolor{ReMAPSecond}78.61 & 79.42 & 74.63 & \cellcolor{ReMAPSecond}68.75 & 49.24 & 48.50 & 24.10 & 39.27 & 69.67 & 62.00 & 51.01 \\
Pixel Reasoner & 81.15 & 75.00 & 81.58 & 73.50 & 65.38 & 58.50 & 53.15 & 27.90 & 44.83 & \cellcolor{ReMAPFirst}\textbf{71.90} & 71.40 & 48.07 \\
ReGround & 73.30 & 73.73 & 77.00 & 68.13 & 60.63 & 54.65 & 56.12 & 25.40 & 28.01 & 66.93 & 69.30 & 43.21 \\
\midrule
\multicolumn{13}{l}{\textit{Encoded-feature reuse}} \\
ICoT & 56.02 & 68.22 & 33.67 & 66.13 & 65.25 & 24.02 & 20.96 & 26.30 & 29.91 & 35.95 & 27.40 & 36.79 \\
MINT-CoT & 41.36 & 72.84 & 74.83 & 54.00 & 43.00 & 49.79 & 54.04 & 26.80 & 34.05 & 62.48 & 62.10 & 47.98 \\
\midrule
\multicolumn{13}{l}{\textit{Latent representations}} \\
DMLR & 44.50 & 68.52 & 66.75 & 46.00 & 42.38 & 42.32 & 39.81 & 22.00 & 23.74 & 54.90 & 51.40 & 35.87 \\
VaLR & 70.68 & 74.79 & 71.75 & 51.00 & 47.88 & 44.88 & 45.08 & 19.50 & 17.93 & 45.75 & 49.30 & 33.21 \\
V-Reflection & 79.06 & 75.21 & 77.75 & 73.25 & 66.50 & 54.21 & 46.65 & 25.10 & 53.16 & 70.85 & 68.70 & 37.94 \\
VisMem & 60.21 & 76.64 & 82.50 & 56.13 & 47.88 & 56.95 & 57.00 & 21.70 & 41.19 & 68.10 & 61.10 & 45.55 \\
\midrule
SFT & 58.64 & 75.04 & 75.58 & 56.62 & 48.25 & 54.13 & 55.73 & \cellcolor{ReMAPSecond}29.00 & 52.70 & 70.59 & \cellcolor{ReMAPSecond}72.10 & \cellcolor{ReMAPSecond}55.05 \\
Stage I$_{\mathrm{sft}}$ & 76.44 & 77.96 & \cellcolor{ReMAPSecond}83.33 & \cellcolor{ReMAPSecond}75.00 & 68.38 & \cellcolor{ReMAPSecond}58.92 & 57.81 & 27.30 & \cellcolor{ReMAPFirst}\textbf{68.37} & 71.24 & 70.10 & 48.35 \\
\textbf{Ours} & \cellcolor{ReMAPSecond}82.20 & \cellcolor{ReMAPFirst}\textbf{79.28} & \cellcolor{ReMAPFirst}\textbf{83.92} & \cellcolor{ReMAPFirst}\textbf{75.50} & \cellcolor{ReMAPFirst}\textbf{69.13} & \cellcolor{ReMAPFirst}\textbf{60.65} & \cellcolor{ReMAPFirst}\textbf{65.38} & \cellcolor{ReMAPFirst}\textbf{29.30} & \cellcolor{ReMAPSecond}68.00 & \cellcolor{ReMAPSecond}71.76 & \cellcolor{ReMAPFirst}\textbf{73.40} & \cellcolor{ReMAPFirst}\textbf{64.27} \\
\bottomrule
\end{tabular}%
}
\end{table}

%% file: sections/04_experiments.tex
\section{Experiments}
\label{sec:experiments}

We evaluate ReMAP on single-image perception, multi-image reasoning, and
broader visual understanding. We then examine the contributions of memory
curation, organization, and access through ablations, latent visualizations,
reasoning trajectories, and visual-context cost.

\subsection{Experimental Setup}
\label{sec:exp-setup}

\afterpage{\input{tables/cross_backbone}}

\paragraph{Benchmarks.}
We use ten benchmark families in three categories.
\textbf{Single-image perception} includes V*Bench~\citep{wu2024vstar},
CV-Bench-2D/3D~\citep{tong2024cambrian}, and
HR-Bench-4K/8K~\citep{wang2025hrbench}.
\textbf{Multi-image reasoning} includes BLINK~\citep{fu2024blink},
MuirBench~\citep{wang2025muirbench}, MMSI-Bench~\citep{yang2026mmsi}, and
MIMIC~\citep{das2026moreimages}.
\textbf{Broader visual understanding}~\citep{yu2024mmvet,lu2024mathvista,xai2024grok15v}
uses MM-Vet, MathVista, and RealWorldQA for generalization tests.

\paragraph{Baselines and implementation.}
Following the approaches discussed in the introduction, we compare
\textbf{pixel reinspection} methods Q-Zoom~\citep{shi2026qzoom},
DeepEyes~\citep{zheng2026deepeyes}, Pixel Reasoner~\citep{su2025pixelreasoner},
and ReGround~\citep{peng2026reground}, \textbf{encoded-feature reuse} methods
ICoT~\citep{gao2025interleaved} and MINT-CoT~\citep{chen2025mintcot}, and
\textbf{latent-representation} methods DMLR~\citep{liu2026reasoning},
VaLR~\citep{jeon2026vision}, V-Reflection~\citep{zhou2026vreflection}, and
VisMem~\citep{yu2026vismem}. Native and Coarse provide backbone controls at
native resolution and at most 1{,}024 initial visual tokens per image.
ReMAP uses the Coarse input budget and returns 32 latents per Local call.
All ReMAP rows and the Native and Coarse controls in
Table~\ref{tab:main-results} use Qwen2.5-VL-7B-Instruct~\citep{bai2025qwen25vl};
prior methods keep their released backbones. ReMAP is trained primarily on 16
NVIDIA A800 GPUs. Evaluation protocols, baseline
configurations, and training details appear in
Appendices~\ref{app:eval-splits}, \ref{app:training-configuration}, and
\ref{app:training-data}.

\subsection{Overall Performance}
\label{sec:exp-main-results}

\ding{182} \textbf{ReMAP outperforms prior methods on all four multi-image benchmarks.}
In Table~\ref{tab:main-results}, ReMAP surpasses the best prior results
on MuirBench and MIMIC by 8.38 and 14.84 percentage points, respectively.
These gains span cross-image comparison, spatial reasoning, and evidence
aggregation. ReMAP also improves over the native-resolution backbone on
all five single-image metrics, extending its benefits to fine-grained
visual perception.

\ding{183} \textbf{ReMAP's learned memory access transfers across backbone families.}
We isolate the contribution of reasoning-time memory invocation by comparing
each backbone's original Base model, its trained Memory-OFF counterpart, and
ReMAP with memory enabled (Table~\ref{tab:cross-backbone}).
ReMAP improves over Memory-OFF on both single-image and multi-image metrics
for all four backbones, showing that the gains come from dynamic evidence
access during reasoning rather than from the trained model weights alone.

\input{sections/04_memory_organization}

\subsection{Memory Access and the Perceptual Cycle}
\label{sec:exp-access}
\label{sec:exp-evidence}

\setcounter{figure}{4}
\input{figures/experiment-analysis/visual_evidence_acquisition}
\setcounter{figure}{4}
\setcounter{figure}{5}
\input{figures/memory-invocation/memory_invocation}
\setcounter{figure}{5}

\textbf{Local provides fine-grained visual evidence, Global provides broader
scene and cross-image context, and their interleaved calls restore the
perceptual cycle.} We analyze how the learned policy routes between these two
forms of visual evidence as reasoning evolves. Figure~\ref{fig:visual-evidence-acquisition}
shows that memory returns are inserted into ongoing reasoning and change the
subsequent reasoning state, yielding a state-dependent evidence--reasoning loop
rather than a fixed access schedule. Figure~\ref{fig:memory-invocation}(b)
shows that Local calls concentrate early and in the middle of the trajectory,
whereas Global calls persist later.

\input{sections/04_ablation}
\finishwrap
\FloatBarrier

\subsection{Visual-Context Cost}
\label{sec:exp-efficiency}

\setcounter{figure}{7}

\textbf{ReMAP maintains strong performance with substantially lower visual-context cost.}
We analyze the accuracy--context-cost tradeoff using the matched profiles in
Figure~\ref{fig:memory-invocation}(a). ReMAP remains competitive while using
fewer visual tokens than Native, and its compact memory returns refresh visual
evidence without repeatedly expanding the decoder context. Detailed accounting
is provided in Appendix~\ref{app:visual-context-protocol}.

%% file: tables/cross_backbone.tex
\begin{table}[!t]
\centering
\footnotesize
\setlength{\tabcolsep}{5pt}
\setlength{\abovecaptionskip}{0pt}\setlength{\belowcaptionskip}{4pt}
\caption{\textbf{Memory access across backbone families.} Base denotes the original backbone, and OFF denotes the trained ReMAP model with memory invocation disabled. $\Delta$ reports ReMAP$-$Base differences in score points. Evaluation details appear in Appendix~\ref{app:cross-backbone-protocol}.}
\label{tab:cross-backbone}
\renewcommand{\arraystretch}{1.18}
\edef\backbonerowheight{\the\dimexpr\arraystretch\ht\strutbox+\arraystretch\dp\strutbox\relax}
\edef\backbonerowdepth{\the\dimexpr\arraystretch\dp\strutbox\relax}
\edef\backboneoffset{\the\dimexpr-\backbonerowheight*3/2\relax}
\sbox0{\fontsize{7.5}{9}\selectfont LLaVA-OneVision-7B}
\edef\backboneblankwidth{\the\dimexpr\wd0+7pt\relax}
\edef\backbonetablewidth{\the\dimexpr\linewidth+23pt\relax}
\resizebox{\linewidth}{!}{%
\begin{tabular*}{\backbonetablewidth}{@{\extracolsep{\fill}}l@{\hspace{7pt}}l*{7}{c}@{}}
\toprule
Backbone & Method & V$^\star$ & CV-3D & HR-8K & BLINK & MuirBench & MMSI & MIMIC Avg. \\
\midrule
\raisebox{\backboneoffset}[0pt][0pt]{\fontsize{7.5}{9}\selectfont Qwen2.5-VL-3B} & Base & 51.83 & 50.83 & 42.12 & 43.61 & 50.08 & 29.40 & 28.21 \\
 & OFF & 56.02 & 70.58 & 49.75 & 46.66 & 50.77 & 28.70 & 36.25 \\
 & ReMAP & 58.64 & 75.58 & 51.25 & 48.45 & 55.23 & 30.50 & 50.11 \\
\rlap{\hspace{\backboneblankwidth}\color{ReMAPSecond}\rule[-\backbonerowdepth]{\dimexpr\backbonetablewidth-\backboneblankwidth\relax}{\backbonerowheight}} & $\Delta$ & {\scriptsize$\textcolor[HTML]{228B22}{\boldsymbol{\uparrow}}\!\mathbf{+6.81}$} & {\scriptsize$\textcolor[HTML]{228B22}{\boldsymbol{\uparrow}}\!\mathbf{+24.75}$} & {\scriptsize$\textcolor[HTML]{228B22}{\boldsymbol{\uparrow}}\!\mathbf{+9.12}$} & {\scriptsize$\textcolor[HTML]{228B22}{\boldsymbol{\uparrow}}\!\mathbf{+4.84}$} & {\scriptsize$\textcolor[HTML]{228B22}{\boldsymbol{\uparrow}}\!\mathbf{+5.15}$} & {\scriptsize$\textcolor[HTML]{228B22}{\boldsymbol{\uparrow}}\!\mathbf{+1.10}$} & {\scriptsize$\textcolor[HTML]{228B22}{\boldsymbol{\uparrow}}\!\mathbf{+21.90}$} \\
\midrule
\raisebox{\backboneoffset}[0pt][0pt]{\fontsize{7.5}{9}\selectfont LLaVA-OneVision-7B} & Base & 70.68 & 45.08 & 56.62 & 44.66 & 41.85 & 28.90 & 44.05 \\
 & OFF & 71.73 & 70.17 & 56.25 & 51.60 & 41.65 & 27.50 & 52.79 \\
 & ReMAP & 76.44 & 76.00 & 59.25 & 53.45 & 48.65 & 29.70 & 59.52 \\
\rlap{\hspace{\backboneblankwidth}\color{ReMAPSecond}\rule[-\backbonerowdepth]{\dimexpr\backbonetablewidth-\backboneblankwidth\relax}{\backbonerowheight}} & $\Delta$ & {\scriptsize$\textcolor[HTML]{228B22}{\boldsymbol{\uparrow}}\!\mathbf{+5.76}$} & {\scriptsize$\textcolor[HTML]{228B22}{\boldsymbol{\uparrow}}\!\mathbf{+30.92}$} & {\scriptsize$\textcolor[HTML]{228B22}{\boldsymbol{\uparrow}}\!\mathbf{+2.62}$} & {\scriptsize$\textcolor[HTML]{228B22}{\boldsymbol{\uparrow}}\!\mathbf{+8.78}$} & {\scriptsize$\textcolor[HTML]{228B22}{\boldsymbol{\uparrow}}\!\mathbf{+6.81}$} & {\scriptsize$\textcolor[HTML]{228B22}{\boldsymbol{\uparrow}}\!\mathbf{+0.80}$} & {\scriptsize$\textcolor[HTML]{228B22}{\boldsymbol{\uparrow}}\!\mathbf{+15.47}$} \\
\midrule
\raisebox{\backboneoffset}[0pt][0pt]{\fontsize{7.5}{9}\selectfont Qwen3-VL-8B} & Base & 62.30 & 92.25 & 59.00 & 64.39 & 73.77 & 27.90 & 56.99 \\
 & OFF & 63.87 & 91.50 & 57.12 & 65.33 & 72.69 & 31.60 & 58.70 \\
 & ReMAP & 66.49 & 93.83 & 60.00 & 67.07 & 74.31 & 32.30 & 74.20 \\
\rlap{\hspace{\backboneblankwidth}\color{ReMAPSecond}\rule[-\backbonerowdepth]{\dimexpr\backbonetablewidth-\backboneblankwidth\relax}{\backbonerowheight}} & $\Delta$ & {\scriptsize$\textcolor[HTML]{228B22}{\boldsymbol{\uparrow}}\!\mathbf{+4.19}$} & {\scriptsize$\textcolor[HTML]{228B22}{\boldsymbol{\uparrow}}\!\mathbf{+1.58}$} & {\scriptsize$\textcolor[HTML]{228B22}{\boldsymbol{\uparrow}}\!\mathbf{+1.00}$} & {\scriptsize$\textcolor[HTML]{228B22}{\boldsymbol{\uparrow}}\!\mathbf{+2.68}$} & {\scriptsize$\textcolor[HTML]{228B22}{\boldsymbol{\uparrow}}\!\mathbf{+0.54}$} & {\scriptsize$\textcolor[HTML]{228B22}{\boldsymbol{\uparrow}}\!\mathbf{+4.40}$} & {\scriptsize$\textcolor[HTML]{228B22}{\boldsymbol{\uparrow}}\!\mathbf{+17.21}$} \\
\midrule
\raisebox{\backboneoffset}[0pt][0pt]{\fontsize{7.5}{9}\selectfont InternVL3-8B} & Base & 59.16 & 68.67 & 62.88 & 53.34 & 51.81 & 27.30 & 36.94 \\
 & OFF & 61.26 & 68.33 & 63.12 & 53.81 & 51.42 & 29.20 & 43.00 \\
 & ReMAP & 65.45 & 69.25 & 64.12 & 55.44 & 52.88 & 30.30 & 57.00 \\
\rlap{\hspace{\backboneblankwidth}\color{ReMAPSecond}\rule[-\backbonerowdepth]{\dimexpr\backbonetablewidth-\backboneblankwidth\relax}{\backbonerowheight}} & $\Delta$ & {\scriptsize$\textcolor[HTML]{228B22}{\boldsymbol{\uparrow}}\!\mathbf{+6.28}$} & {\scriptsize$\textcolor[HTML]{228B22}{\boldsymbol{\uparrow}}\!\mathbf{+0.58}$} & {\scriptsize$\textcolor[HTML]{228B22}{\boldsymbol{\uparrow}}\!\mathbf{+1.25}$} & {\scriptsize$\textcolor[HTML]{228B22}{\boldsymbol{\uparrow}}\!\mathbf{+2.10}$} & {\scriptsize$\textcolor[HTML]{228B22}{\boldsymbol{\uparrow}}\!\mathbf{+1.08}$} & {\scriptsize$\textcolor[HTML]{228B22}{\boldsymbol{\uparrow}}\!\mathbf{+3.00}$} & {\scriptsize$\textcolor[HTML]{228B22}{\boldsymbol{\uparrow}}\!\mathbf{+20.06}$} \\
\bottomrule
\end{tabular*}
}%
\end{table}

%% file: sections/04_memory_organization.tex
\subsection{Interpreting Latent Memory Organization}
\label{sec:exp-memory-organization}

\setcounter{figure}{3}
\input{figures/latent-memory/memory_organization}
\setcounter{figure}{3}

\textbf{Global and Local form two distinct types of latent memory.} Figure~\ref{fig:memory-organization}
visualizes their benchmark-conditioned t-SNE projections. Local is
benchmark-specific: its distribution changes across benchmarks. Global mixes
across benchmarks because it captures abstract scene-level and cross-image
relations. In the multi-image view, Global is more dispersed, suggesting that
it represents a wider range of inter-image relations.

\finishwrap
\FloatBarrier

%% file: figures/latent-memory/memory_organization.tex
\begin{figure}[t]
  \centering
  \begin{minipage}[t]{0.42\textwidth}
    \centering
    {\small\textbf{(a) High-resolution benchmarks}}\par\vspace{2pt}
    \includegraphics[width=\linewidth,trim=0 0 0 17.2pt,clip]{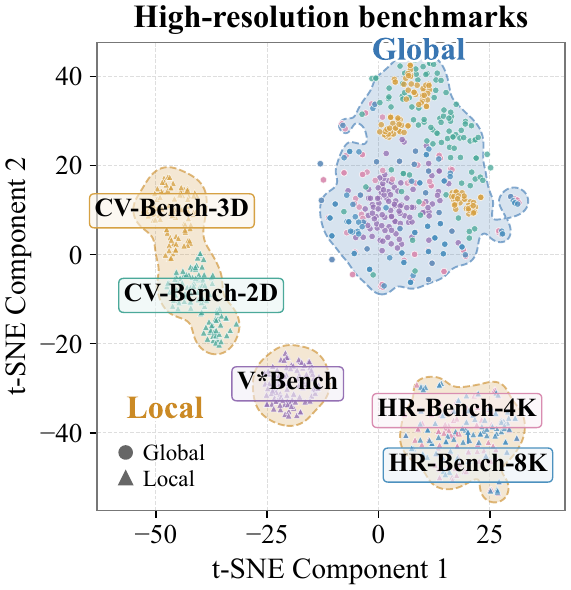}
  \end{minipage}\hspace{0.06\textwidth}%
  \begin{minipage}[t]{0.42\textwidth}
    \centering
    {\small\textbf{(b) Multi-image benchmarks}}\par\vspace{2pt}
    \includegraphics[width=\linewidth,trim=0 0 0 17.2pt,clip]{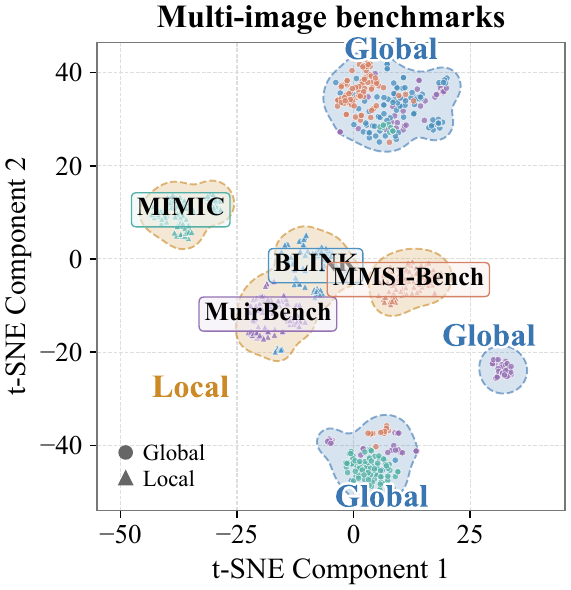}
  \end{minipage}
  \caption{\textbf{Global and Local memory representations.}
  (a) High-resolution and (b) multi-image benchmark groups. Joint t-SNE
  projections show mean-pooled Global and Local memory vectors; colors denote
  benchmark groups, circles and triangles denote Global and Local,
  respectively, and shaded contours show density.}
  \label{fig:memory-organization}
\end{figure}

%% file: figures/experiment-analysis/visual_evidence_acquisition.tex
\begin{figure}[!ht]
  \centering
  \includegraphics[width=\linewidth,trim=0 0 0 14pt,clip]{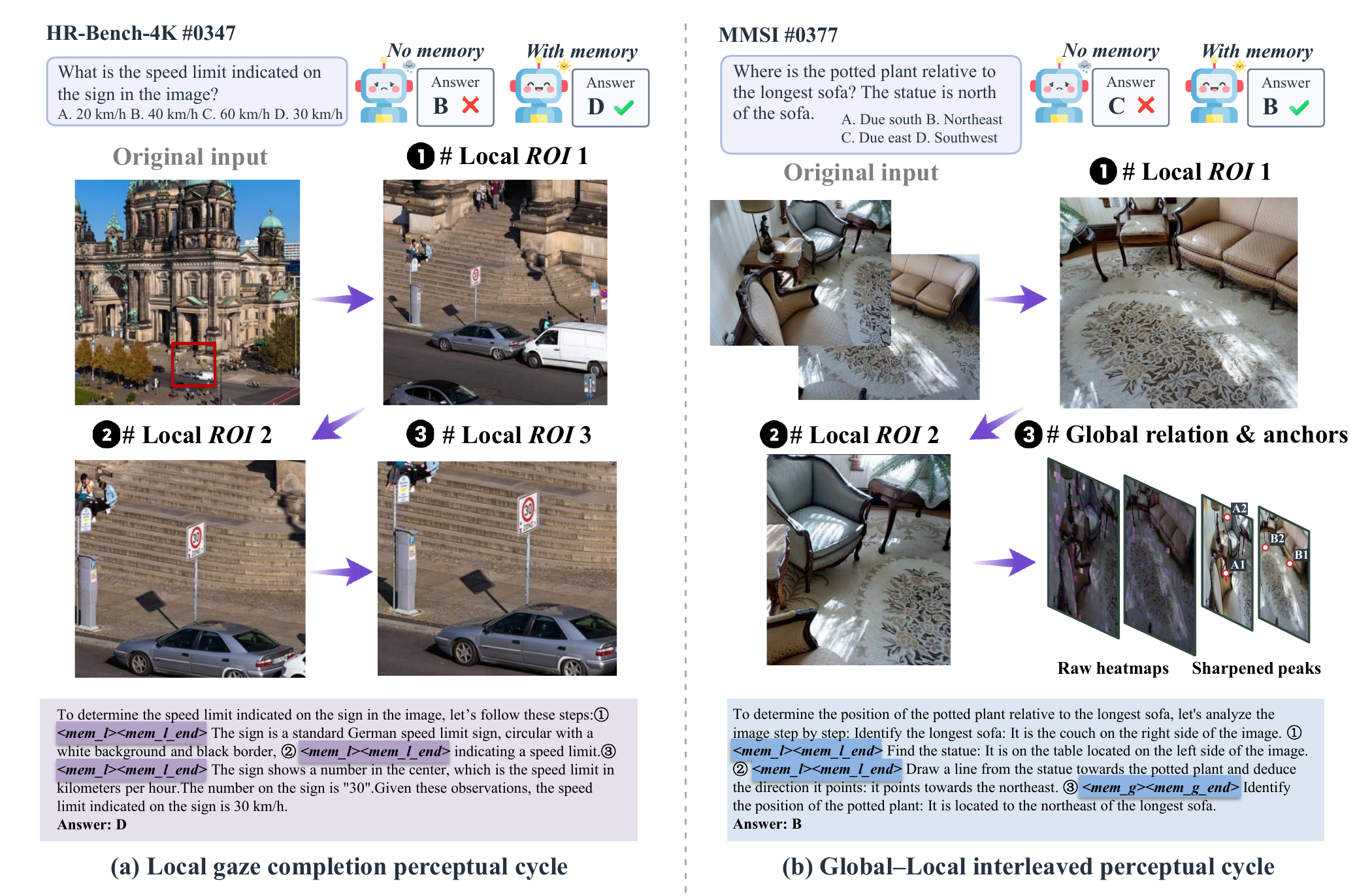}
  \caption{\textbf{Local refinement and Global--Local interaction.}
  (a) HR-Bench-4K \#0347: repeated Local calls narrow the inspection from
  the street scene to a speed-limit sign, yielding 30~km/h rather than
  40~km/h without memory. (b) MMSI \#0377: two Local calls followed by a
  Global call accompany the northeast answer, whereas the paired no-call
  trajectory answers due east.}
  \label{fig:visual-evidence-acquisition}
\end{figure}

%% file: figures/memory-invocation/memory_invocation.tex
\begin{figure*}[!t]
  \centering
  \begin{minipage}[b]{0.625\textwidth}
    \centering
    \includegraphics[width=\linewidth,trim=0 6pt 0 29.5pt,clip]{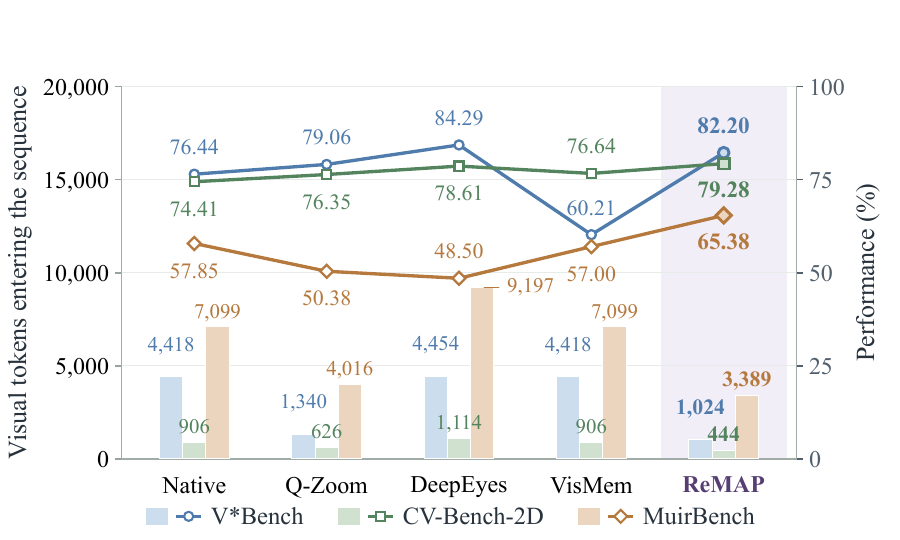}
    \par{\small\textbf{(a)}}
  \end{minipage}\hfill
  \begin{minipage}[b]{0.355\textwidth}
    \centering
    \includegraphics[width=\linewidth,trim=0 7.9pt 0 39pt,clip]{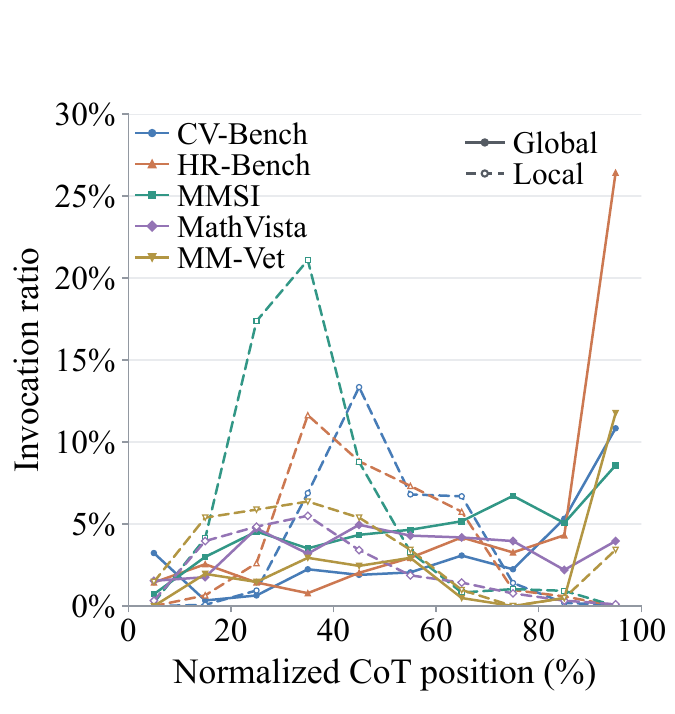}
    \par{\small\textbf{(b)}}
  \end{minipage}
  \caption{\textbf{Memory invocation and visual-context cost.}
  (a) Mean total visual tokens entering the sequence (bars) and performance
  (lines) on shared V*Bench, CV-Bench-2D, and MuirBench questions; the shaded
  column marks ReMAP. (b) Global (solid) and Local (dashed) invocation ratios
  over the normalized reasoning trajectory.}
  \label{fig:memory-invocation}
  \label{fig:actual-visual-context}
\end{figure*}

%% file: sections/04_ablation.tex
\subsection{Component, Capacity, and Access Ablations}
\label{sec:exp-ablation}

Next, we study the gains from different components and design choices in ReMAP,
including training, memory curation, memory access, and latent capacity.
\par

\input{tables/component_ablation}
\setcounter{figure}{6}
\begin{figure}[t]
  \input{figures/experiment-analysis/latent_capacity_ablation}
\end{figure}
\setcounter{figure}{6}

\textbf{Global and Local specialize by scope, and their combination is best.}
We analyze how curation scope affects benchmark performance. Global is stronger
on CV-Bench-3D, MuirBench, and MM-Vet, whereas Local is stronger on
V$^\star$Bench and BLINK. Combining both memories is best on every benchmark,
exceeding the better of the Global-only and Local-only variants by 3.72 points
on MM-Vet, as shown in Table~\ref{tab:component-ablation}.

\textbf{Learned invocation is more reliable than fixed or heuristic schedules.}
We analyze whether the learned access policy selects more useful request points
than Fixed, Random, and Entropy schedules. ReMAP is highest on every benchmark,
exceeding the strongest of these alternatives by 3.38 points on MuirBench and
4.64 on MM-Vet. Fixed invocation falls below w/o Memory on V$^\star$Bench,
showing that a fixed schedule can make added evidence harmful; Table~\ref{tab:component-ablation}
reports the access variants. The control definitions are given in
Appendix~\ref{app:additional-ablations}. Appendix~\ref{app:branch-selection-diagnostic}
examines branch-point selection and routing diversity.

\textbf{ReMAP remains effective across a range of latent return sizes, making
a compact capacity sufficient.} We sweep the Local and Global return sizes to
determine the default configuration (Figure~\ref{fig:latent-capacity-ablation}).
Performance is broadly stable across the tested capacities, while increasing
the number of returned tokens does not bring consistent gains. We therefore use
$K=32$ Local tokens and $4N+4$ Global tokens in the remaining experiments.

%% file: tables/component_ablation.tex
\begin{table}[t]
\centering
\caption{\textbf{Training, curation, and access ablations (\%).} Base is Coarse; BLINK reports micro accuracy and MM-Vet mean graded scores.}
\label{tab:component-ablation}
\small
\begin{tabular*}{0.74\textwidth}{@{\extracolsep{\fill}}lrrrrr@{}}
\toprule
Method & V$^\star$ & CV-3D & Muir & BLINK & MM-Vet \\
\midrule
Base & 67.54 & 75.42 & 58.15 & 57.81 & 47.06 \\
SFT & 58.64 & 75.58 & 55.73 & 54.13 & 55.05 \\
Stage I$_{\mathrm{sft}}$ & 76.44 & 83.33 & 57.81 & 58.92 & 48.35 \\
\addlinespace[2pt]
\multicolumn{6}{@{}c@{}}{\color[HTML]{2459A6}\null\leaders\hrule height 2.1pt depth -1.7pt\hskip 0pt plus 1filll\relax\hspace{0.7em}\textit{Memory curation}\hspace{0.7em}\leaders\hrule height 2.1pt depth -1.7pt\hskip 0pt plus 1filll\relax\null} \\
\addlinespace[1pt]
w/o Memory & 79.06 & 80.00 & 61.00 & 59.02 & 57.80 \\
w/ Global memory & 79.58 & 83.00 & 64.00 & 59.50 & 60.55 \\
w/ Local memory & 81.68 & 81.00 & 62.00 & 60.02 & 59.63 \\
\addlinespace[2pt]
\multicolumn{6}{@{}c@{}}{\color[HTML]{2459A6}\null\leaders\hrule height 2.1pt depth -1.7pt\hskip 0pt plus 1filll\relax\hspace{0.7em}\textit{Memory access}\hspace{0.7em}\leaders\hrule height 2.1pt depth -1.7pt\hskip 0pt plus 1filll\relax\null} \\
\addlinespace[1pt]
Fixed & 76.96 & 83.00 & 61.46 & 59.39 & 58.72 \\
Random & 79.58 & 79.75 & 61.58 & 58.81 & 57.34 \\
Entropy & 79.06 & 81.17 & 62.00 & 59.39 & 59.63 \\
\midrule
\textbf{ReMAP} & \textbf{82.20} & \textbf{83.92} & \textbf{65.38} & \textbf{60.65} & \textbf{64.27} \\
\bottomrule
\end{tabular*}
\end{table}

%% file: figures/experiment-analysis/latent_capacity_ablation.tex
\centering
\includegraphics[width=0.72\linewidth]{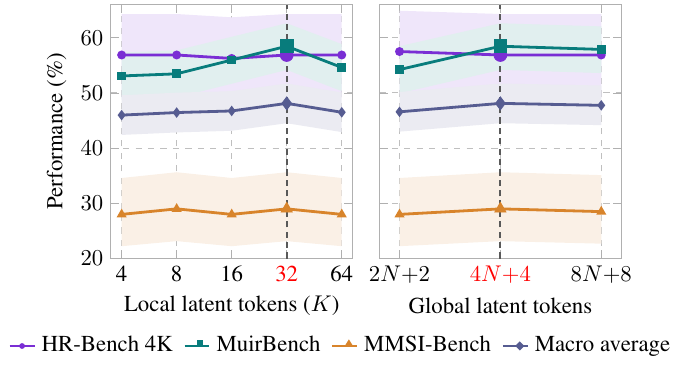}
\caption{\textbf{Latent-capacity sensitivity.}
Performance on 20\% holdout splits from HR-Bench-4K, MuirBench, and MMSI-Bench.
Curves show benchmark scores and
their equal-weight macro average; shaded bands are item-level 95\% intervals
from one training seed. Dashed guides mark the selected $K=32$ and $4N+4$
settings.}
\label{fig:latent-capacity-ablation}

%% file: sections/05_limitations.tex
\section{Limitations}
\label{sec:limitations}

We note three limitations of ReMAP that we hope to address in future work.
First, ReMAP operates within a fixed memory and access budget: each
trajectory permits at most five memory requests, each Local request
re-encodes at most three selected images, and each Global block has a fixed
$4N+4$ layout. These limits keep computation low but may omit evidence when an
input contains many images or when a task requires extended visual search.
Second, Twig is trained only in Stage~I on teacher-reviewed ROI targets and
receives no further training dedicated to localization. When its box misses
the relevant region, the subsequent zoom-in contributes less useful detail,
so we expect a stronger localizer, for example one trained with dedicated
grounding data, to further improve Local memory. Third, the memory modules
are attached at fixed positions of the backbone: the Source Selector reads
layer 12, and Twig reads layer 18. Internal layer sweeps selected these taps
from the quality of the tapped representations, but we have not yet studied
how placing the components at other layers affects end-to-end performance,
and we leave this analysis to future work.

%% file: sections/05_conclusion.tex
\Needspace{5\baselineskip}
\section{Conclusion}
\label{sec:conclusion}

ReMAP shows that visual memory becomes effective when the scope and timing of
visual evidence follow the reasoning state. Global memory preserves
question-conditioned scene and cross-image context, while Local memory
reacquires fine-grained pixels as the trajectory demands. A learned policy
coordinates these two forms of evidence within compact latent returns,
improving over prior methods on all four multi-image benchmarks and over the
Native backbone on all five single-image metrics while reducing the profiled
visual context. Together, these components restore the perceptual cycle by
letting each reasoning step use its current evidence to guide the next visual
inspection. We hope this view helps future multimodal models treat
perception as part of reasoning.

%% file: statements/ai_use.tex
We used generative AI tools to aid and polish the writing, retrieve and
discover related literature, and draft sections of the manuscript. We
reviewed all AI-assisted material and take responsibility for the final
content of this paper.

%% file: statements/ethics.tex
This work uses public research datasets and no human subjects or newly
collected personal data. ReMAP does not remove inherited model biases or
failure modes; additional visual operations can also make source-allocation
or localization errors appear confident.

%% file: statements/reproducibility.tex
Details of the memory interface, learning objectives, training
configuration, data, evaluation protocols, and ablations are provided in
the appendices. We will publicly release our source code and model weights.

%% file: sections/90_appendix.tex
\section{Appendix}
\label{app:supplement}

\input{sections/92_related_work}

\subsection{Runtime and Memory Interface}
\label{app:runtime}

\paragraph{Memory transactions.}
ReMAP begins autoregressive reasoning from a coarse visual prefix of the input
images. Ordinary autoregressive generation continues by default. A memory
transaction is performed only when the decoder emits a Global or Local request
token. The corresponding transactions are
\begin{equation}
  \big[\langle\texttt{mem\_g}\rangle;\,Z_G;\,
  \langle\texttt{mem\_g\_end}\rangle\big],
  \qquad
  \big[\langle\texttt{mem\_l}\rangle;\,Z_L;\,
  \langle\texttt{mem\_l\_end}\rangle\big].
  \label{eq:memory-transaction}
\end{equation}
After the decoder emits a request token, the runtime inserts the corresponding
latent block and end marker into the current decoder sequence. Generation
resumes from the position following the end marker. End markers are not
sampled as vocabulary outputs, and returned latents enter the decoder as
continuous inputs. Each reasoning trajectory permits at most five memory
requests. After the request budget is exhausted or the model commits an
answer, Global and Local request tokens are masked.

Inserted latents and end markers occupy successive positions in the
decoder sequence and KV cache. Request tokens and these inserted positions
do not count toward the ordinary-text budget, but they still consume decoder
context capacity.

\paragraph{Global memory.}
Global curation forms a question-conditioned latent block from the ordered
input images. For an input containing $N$ images, four tokens are allocated to
each image and four additional tokens encode cross-image relations:
\begin{equation}
  K_G(N)=4N+4.
\end{equation}
The resulting block is formed before decoding, cached throughout the reasoning
trajectory, and inserted at a new sequence position whenever the decoder emits
a Global request. Repeated Global requests neither re-encode the input images
nor regenerate the cached block from a later reasoning state.

\paragraph{Local memory.}
The Local request uses the current reasoning state to select image sources and
regions. The selector retains at most three sources, stopping when cumulative
selection probability reaches 0.85. A frozen vision encoder maps the selected
crops to region features, and the Local Weaver combines these features with
the reasoning state and cached Global memory:
\begin{equation}
  Z_L=\left[W_L(h_t,R_{t,i};Z_G)\right]_{i\in\mathcal S_t}.
\end{equation}
Each Local request returns 32 latent tokens. Global memory supplies context
for Local compression but is not concatenated into the 32-token Local output.

The $4N+4$ Global layout, the 32-token Local return, the three-source cap,
and the 0.85 threshold define the memory interface used in all main
experiments.
The Source Selector reads the layer-12 state exposed at a request boundary,
and Twig uses layer-18 prefix states selected by a layer sweep. Stage~II
rollout branching follows the partial-rollout strategy of ARPO~\citep{dong2025arpo}
and is specified with the access objective below.

\subsection{Memory Formation and Access Objectives}
\label{app:formation-access-objectives}

\subsubsection{Global Layout and Caching}

Each per-image Image-Weaver span contains four latent queries and is restricted
by its attention mask to the visual activations of its own source image. Thus
$Z_i^{\mathrm{img}}\in\mathbb{R}^{4\times d}$. The Relation Weaver uses four
queries over the concatenated image summaries
$[Z_1^{\mathrm{img}};\cdots;Z_N^{\mathrm{img}}]$; its relation span can attend
to all sources, giving $\mathcal Z^{\mathrm{rel}}\in\mathbb{R}^{4\times d}$.
Both weavers are conditioned on the encoded question, as in
Equation~\ref{eq:image-weaver}. The Global block is therefore

\begin{equation}
  Z_G=[Z_1^{\mathrm{img}};\cdots;Z_N^{\mathrm{img}};
  \mathcal Z^{\mathrm{rel}}]\in\mathbb{R}^{(4N+4)\times d},
\end{equation}

which is the $K_G(N)=4N+4$ layout used in the capacity sweep of
Figure~\ref{fig:latent-capacity-ablation}. We compute this block once from the
prompt-time images and question, retain it as $Z_G$, and reuse the same latent
values at every later Global request. Only the decoder insertion position and
its resulting KV entries are new; later reasoning states determine whether a
request is made, but do not recompute or alter the cached block.

\subsubsection{Local Selection and Compression}
\label{app:local-formation}

The Source Selector uses the layer-12 request state $h_t=h_t^{12}$ to rank
the source images. If $p_{t,i}$ denotes the normalized score of source $i$,
the selector orders sources by decreasing score and keeps the shortest prefix
whose cumulative score reaches 0.85, with a maximum of three sources. When the
threshold is not reached within three sources, it keeps the three highest-
scoring sources. The layer-12 request tap and layer-18 Twig tap are ReMAP
settings selected by internal layer sweeps: layer 12 supplies the state at a
request boundary, while layer 18 provides the spatially arranged activations
used for localization.
Twig Localization uses the layer-18 prefix states $H_{<t}^{18}$ to locate a
region $b_{t,i}$ in each selected image $I_i$.
Let $V_i^c$ denote the coarse visual features of image $I_i$. Source selection,
region localization, and frozen crop encoding are written as
\begin{equation}
  \begin{aligned}
    \mathcal S_t&=f_{\mathrm{sel}}\big(h_t,\{V_i^{c}\}_{i=1}^{N}\big),\\
    b_{t,i}&=f_{\mathrm{loc}}\big(H_{<t}^{18},V_i^{c}\big),\qquad
    R_{t,i}=E_{\mathrm{roi}}\big(I_i[b_{t,i}]\big).
  \end{aligned}
  \label{eq:localize}
\end{equation}
Here, $I_i[b_{t,i}]$ is the selected crop, and $E_{\mathrm{roi}}$ is the frozen
vision encoder that produces its region features $R_{t,i}$.

After processing $H_{<t}^{18}$ with its decoder blocks, Twig applies
query/key projections and multimodal rotary position embeddings (mRoPE).
In Figure~\ref{fig:local-memory-details}(a), $q_t^h$ denotes the query at the
current reasoning position for head $h$, and $K_{\mathrm{img}}^h$ contains the
keys at the selected image's visual-token positions. The ROI heatmap is the
head-averaged scaled dot product reshaped to the image's coarse spatial grid:
\begin{equation}
  q_{\mathrm{roi}}=\operatorname{reshape}_{G_h\times G_w}
  \left(\frac{1}{H}\sum_{h=1}^{H}
  \frac{q_t^h(K_{\mathrm{img}}^h)^{\top}}{\sqrt{d_h}}\right).
  \label{eq:roi-heatmap}
\end{equation}
For readout, $q_{\mathrm{roi}}$ is treated as a logit map consistent with the
weighted-BCE localization objective. We retain grid cells with
$q_{\mathrm{roi}}>0$ (equivalently, sigmoid probability at least $0.5$), take
the min--max box of their row and column coordinates, map the box from the
coarse grid to image pixels, and clip it to the image boundary. The resulting
box is $b_{t,i}$ for crop encoding.

For local compression, the Local Weaver conditions its learnable compression
queries $Q_L$ on the reasoning state $h_t$ and cached Global memory $Z_G$.
Twig's $q_t$ is used only for region localization. The crop features $R_{t,i}$
provide fine-grained evidence, while $Z_G$ supplies image summaries and
cross-image relational context. Equation~\ref{eq:local-memory} summarizes
source-specific compression and concatenation into a return with $K=32$
latent tokens in the default configuration.

\subsubsection{Formation and Warm-up Objectives}
\label{app:formation-objectives}

\paragraph{Continuation distillation.}
The Image, Relation, and Local Weavers are trained through the frozen LLM.
For a teacher trajectory with prefix $\mathcal P_t$ and continuation $Y$, we
insert a Global or Local transaction at position $t$ and minimize
\begin{equation}
  \mathcal L_{\mathrm{suffix}}
  =-\frac{1}{|Y|}\sum_{k=1}^{|Y|}
    \log p_{\mathrm{LLM}}\big(Y_k\mid\mathcal P_t,\,
    [\langle\texttt{mem}\rangle;Z;\langle\texttt{mem\_end}\rangle],\,
    Y_{<k}\big),
  \label{eq:suffix-loss}
\end{equation}
where $Z$ is $Z_G$ or $Z_L$ and the markers match its memory type. The LLM
is frozen, so gradients reach only the weaver that produced $Z$. The loss
therefore trains each weaver to encode the evidence that the teacher's
subsequent reasoning uses. Together with $\mathcal L_{\mathrm{suffix}}$,
fact, relation, and answer supervision from the Stage~I annotations
(Appendix~\ref{app:training-data}) trains the weavers.

\paragraph{ROI localization.}
Twig produces ROI logits $q_{\mathrm{roi}}$ on the coarse grid of each
selected image. The teacher-reviewed box is rasterized to a binary target
$\bar A$ on the same grid. With $P=\operatorname{sigmoid}(q_{\mathrm{roi}})$,
the localization objective is
\begin{equation}
  \begin{aligned}
  \mathcal L_{\mathrm{ROI}}
  &=\mathcal L_{\mathrm{wBCE}}(\bar A,q_{\mathrm{roi}})
  +1-\operatorname{softIoU}(\bar A,P),\\
  \operatorname{softIoU}(\bar A,P)
  &=\frac{\sum\bar A\odot P}{\sum\bar A+\sum P-\sum\bar A\odot P}.
  \end{aligned}
  \label{eq:local-supervision}
\end{equation}
Q-Zoom~\citep{shi2026qzoom} trains its ROI predictor with binary
cross-entropy. We weight positive cells by the ratio of negative to positive
cells, so that a small target region is not dominated by background cells. We cap this weight at 50 based on our experience. The
soft-IoU term matches the overall extent of the heatmap to the target box.

\paragraph{Student-state adaptation.}
Teacher trajectories do not contain the states that the student reaches at
inference. We therefore run the student autoregressively and apply the same
two objectives to its states. For a student state $s_t$ with complete
text-and-latent prefix $\mathcal P_t$, the teacher reviews $\mathcal P_t$
together with the selected crop $I_i[b_{t,i}]$ and returns an ROI target
$\bar A$ and a continuation target $Y$. Twig is updated with
$\mathcal L_{\mathrm{ROI}}$, and the weavers with $\mathcal L_{\mathrm{suffix}}$
conditioned on $\mathcal P_t$.

\paragraph{Interface warm-up.}
Each warm-up sequence either contains no memory transaction or contains one
Global or Local transaction at a request boundary. The frozen formation
modules compute $Z$, which enters the sequence as a constant input. Warm-up
minimizes next-token cross-entropy over the supervised positions $\mathcal T$:
\begin{equation}
  \mathcal L_{\mathrm{warm}}
  =-\frac{1}{|\mathcal T|}\sum_{k\in\mathcal T}
    \log\pi(y_k\mid y_{<k}),
  \qquad
  \mathcal T=\mathcal T_{\mathrm{call}}\cup\mathcal T_{\mathrm{end}}
  \cup\mathcal T_{\mathrm{later}}.
  \label{eq:warmup-loss}
\end{equation}
Here $\mathcal T_{\mathrm{call}}$ contains the request token,
$\mathcal T_{\mathrm{end}}$ the first text token after the end marker, and
$\mathcal T_{\mathrm{later}}$ the remaining reasoning and answer tokens. The
three sets give $\mathcal L_{\mathrm{call}}$, $\mathcal L_{\mathrm{end}}$,
and $\mathcal L_{\mathrm{later}}$ in
Figure~\ref{fig:local-memory-details}(b). Prompt tokens, inserted latents,
and end markers are excluded from $\mathcal T$. Sequences without a
transaction supervise ordinary text at states where no memory is requested,
so the decoder does not learn to request memory at every state. Only the four
marker input rows and the two request output rows are updated. The end-marker
input rows receive gradients through the text predicted after the return.

\subsubsection{Access Learning Objective}
\label{app:access-objectives}

\paragraph{Policy parameters.}
Let $\pi_{\phi,\psi}$ denote the decoding policy. The symbol $\phi$ denotes
the decoder LoRA parameters, and
$\psi=(\eta_{\mathrm G},\eta_{\mathrm L},w_{\mathrm G},w_{\mathrm L})$
denotes the request interface, where $\eta_{\mathrm G}$ and
$\eta_{\mathrm L}$ are the request-token input embeddings and
$w_{\mathrm G}$ and $w_{\mathrm L}$ are their output-classifier rows. All other parameters, including the end-marker
embeddings and the memory-formation modules, stay fixed. Let
$\mathcal V_t$ be the legal vocabulary and $h_{\phi,\eta}(\mathcal P_t)$ the
decoder state over the complete text-and-latent prefix. The request rows
share the output softmax with the frozen ordinary rows $w_v$:
\begin{equation}
  \ell(v\mid s_t)=w_v^\top h_{\phi,\eta}(\mathcal P_t),
  \qquad
  \pi_{\phi,\psi}(v\mid s_t)=
  \frac{\exp\ell(v\mid s_t)}
       {\sum_{u\in\mathcal V_t}\exp\ell(u\mid s_t)}.
  \label{eq:request-head}
\end{equation}
This softmax gives $p_t(\mathrm G)$ and $p_t(\mathrm L)$, and the remaining
mass is $p_t(\mathrm N)=1-p_t(\mathrm G)-p_t(\mathrm L)$. An ordinary
output $z$ has probability
$p_{\phi,\psi}(\mathrm N\mid s_t)\,q_{\phi,\eta}(z\mid s_t,\mathrm N)$.
Updating $w_{\mathrm G}$ and $w_{\mathrm L}$ changes the request logits and
the shared normalizer, and updating $\eta_{\mathrm G}$ and $\eta_{\mathrm L}$
changes the hidden states after a request. Invocation credit therefore acts
on the whole suffix after a request, including the text generated after the
returned latents.

\paragraph{Rollouts and branch selection.}
Each prompt samples four root trajectories from the behavior policy
$\pi_{\mathrm{old}}$. Candidate branch states precede answer commitment and
admit at least one legal request, including the initial state. Each root
branches once, at the candidate with the largest $u(s_t,a_t)$ in
Equation~\ref{eq:crd}, with ties resolved by the earliest position. From the shared
prefix, two siblings ($m=2$) are sampled independently, each with its own
subsequent requests and reasoning. Requests in the shared prefix count toward
the five-request budget. The roots and their siblings form the natural group
$\mathcal G$, so $4\leq|\mathcal G|\leq12$.

\paragraph{Rewards.}
For a completed trajectory $\tau_i$, the terminal reward scores its extracted
answer $\hat y(\tau_i)$ against the reference $y^\star$:
\begin{equation}
  R_i=\operatorname{Score}\!\left(\hat y(\tau_i),y^\star\right)\in[0,1].
  \label{eq:task-reward}
\end{equation}
Typed answer checks are used where applicable, and semantic evaluation gives
graded scores; for example, listing questions receive per-example F1. A
prompt enters the policy losses only when all natural trajectories and the
required continuation controls have valid scores. Unavailable scores are not
treated as zero.

\paragraph{Same-state continuation baseline.}
For branch $b$, the siblings at $b$ are the root and its resampled
continuations, and $\mathcal G_b$ contains them. Let
$D_b=\{i\in\mathcal G_b:a_{i,b}=\mathrm N\}$ contain the siblings that
continue without a request at $b$, which can include the root. The baseline
$B_b$ in Section~\ref{sec:method-learning} is their mean reward:
\begin{equation}
  B_b=
  \begin{cases}
    |D_b|^{-1}\sum_{i\in D_b}R_i, & D_b\neq\varnothing,\\
    R(\tau_b^{\mathrm{aux}}), & D_b=\varnothing,
  \end{cases}
  \label{eq:continuation-baseline}
\end{equation}
where $\tau_b^{\mathrm{aux}}$ starts from the same branch state and is forced
to emit an ordinary token at its first decision. Its later decisions follow
$\pi_{\mathrm{old}}$ and may request memory. The auxiliary continuation only
supplies $B_b$ and does not enter $\mathcal G$.

\paragraph{Advantages.}
The decoder uses the group-relative advantage of
GRPO~\citep{shao2024deepseekmath}, and the request interface uses the
normalized gain over continuation at the same branch state:
\begin{equation}
  \hat A_i^{\mathrm{dec}}
  =\frac{R_i-\bar R_{\mathcal G}}{\sigma_{\mathcal G}+\delta},
  \qquad
  \hat A_i^{\mathrm{call}}
  =\frac{R_i-B_{b(i)}}{\tilde\sigma_{b(i)}},
  \qquad
  \tilde\sigma_b=
  \begin{cases}
    \sigma_b+\delta, & \sigma_b>\delta,\\
    1, & \sigma_b\leq\delta.
  \end{cases}
  \label{eq:access-advantages}
\end{equation}
Here $\bar R_{\mathcal G}$ and $\sigma_{\mathcal G}$ are the mean and sample
standard deviation of rewards in $\mathcal G$, and $b(i)$ is the branch at
which trajectory $i$ requests memory. The branch scale $\sigma_b$ is the
sample standard deviation of the rewards in $\mathcal G_b$ together with one
additional entry $B_b$. We set $\delta=10^{-6}$. Equal rewards give zero
advantage in both terms.

\paragraph{Policy objectives.}
Let $z_{ij}$ be the vocabulary output at decision $j$ of trajectory $i$,
including ordinary text, request tokens, and EOS. Its likelihood ratio
$\rho_{ij}=\pi_{\phi,\psi}(z_{ij}\mid s_{ij})/
\pi_{\mathrm{old}}(z_{ij}\mid s_{ij})$ uses the same legal-output mask as
sampling. We use the dual-clipped policy loss~\citep{ye2020mastering}:
\begin{equation}
  \begin{aligned}
  \ell_{\mathrm{clip}}(\rho,A)
    &=-\min\!\left\{\rho A,\,
      \operatorname{clip}(\rho,1-\epsilon,1+\epsilon)A\right\},\\
  \ell_{\mathrm{dc}}(\rho,A)
    &=
    \begin{cases}
      \min\{\ell_{\mathrm{clip}}(\rho,A),-cA\}, & A<0,\\
      \ell_{\mathrm{clip}}(\rho,A), & A\geq0,
    \end{cases}
  \end{aligned}
  \label{eq:clipped-access-loss}
\end{equation}
where $\epsilon$ is the ratio-clipping width and $c>1$ bounds the loss for
negative advantages. The decoder term uses all $T_i$ sampled decisions of each
natural trajectory, including request tokens, EOS, and the shared prefix of
each sibling, with $N_{\mathrm{dec}}=\sum_{i\in\mathcal G}T_i$. The invocation
term uses the set $\mathcal C$ of decisions from natural trajectories that
request memory at their branch, from the request through answer commitment;
it includes ordinary text as well as later requests. Returned latents and end
markers are inserted by the runtime and contribute no policy term. With
$N_{\mathrm{call}}=\max(1,|\mathcal C|)$, the two objectives are
\begin{equation}
  \begin{aligned}
  \mathcal L_{\mathrm{dec}}
    &=\frac{1}{N_{\mathrm{dec}}}
      \sum_{i\in\mathcal G}\sum_{j=1}^{T_i}
      \ell_{\mathrm{dc}}(\rho_{ij},\hat A_i^{\mathrm{dec}})
      +\lambda_{\mathrm{CE}}\mathcal L_{\mathrm{CE}},\\
  \mathcal L_{\mathrm{access}}
    &=\frac{1}{N_{\mathrm{call}}}
      \sum_{(i,j)\in\mathcal C}
      \ell_{\mathrm{dc}}(\rho_{ij},\hat A_i^{\mathrm{call}})
      +\lambda_{\mathrm{pref}}\mathcal L_{\mathrm{pref}}.
  \end{aligned}
  \label{eq:access-objectives}
\end{equation}
The policy part of $\mathcal L_{\mathrm{access}}$ is zero when $\mathcal C$ is
empty. Only $\phi$ receives gradients from $\mathcal L_{\mathrm{dec}}$, and
only $\psi$ receives gradients from $\mathcal L_{\mathrm{access}}$. Each
objective is computed per prompt and averaged over the prompts eligible for
it. The values of $\epsilon$, $c$, $\lambda_{\mathrm{CE}}$, and
$\lambda_{\mathrm{pref}}$ are listed in
Table~\ref{tab:optimization-settings}.

\paragraph{Post-call answer supervision.}
Let $\mathcal P_t^+$ be the complete prefix just after a memory return,
including the request, its latents, and its end marker. The target $Y$
contains the answer-format prefix, the reference answer, and EOS. With memory
requests masked, the auxiliary loss is
\begin{equation}
  \mathcal L_{\mathrm{CE}}
  =-\frac{1}{|Y|}\sum_{k=1}^{|Y|}
    \log\pi_{\phi,\psi}^{\mathrm{text}}
    (Y_k\mid\mathcal P_t^+,Y_{<k}).
  \label{eq:post-call-ce}
\end{equation}
The return is taken from the last pre-answer request of the first root that
makes one. If no root requests memory, a forced initial Global or Local probe
supplies the return. The prefix $\mathcal P_t^+$ contains no later returns,
reasoning suffix, or reference answer. This term trains the decoder to answer
from returned evidence and needs only a valid reference, independent of the
rollout reward.

\paragraph{Initial-state request preferences.}
Three probes start from the initial state $s_0$. One is forced to emit an
ordinary token, and the other two emit the Global and Local request tokens;
later decisions follow $\pi_{\mathrm{old}}$. Let
$\Delta_a=R(\tau_0^a)-R(\tau_0^{\mathrm N})$ for
$a\in\{\mathrm G,\mathrm L\}$. The preference loss is
\begin{equation}
  \mathcal L_{\mathrm{pref}}
  =\frac12\sum_{a\in\{\mathrm G,\mathrm L\}}|\Delta_a|\,
    \operatorname{softplus}\!\left[
      -\operatorname{sgn}(\Delta_a)
      \log\frac{p_\psi(a\mid s_0)}{p_\psi(\mathrm N\mid s_0)}
    \right].
  \label{eq:initial-state-preference}
\end{equation}
The loss raises the probability of a request relative to continuation when
that request improves the reward, and lowers it otherwise. The initial hidden
state is detached, so only $w_{\mathrm G}$ and $w_{\mathrm L}$ receive
gradients. The probes do not enter either policy term. A prompt uses this term
only when all three probes have valid scores.

\subsection{Training Configuration}
\label{app:training-configuration}

Figure~\ref{fig:app-training-pipeline} shows the five training steps. The
vision encoder and the pretrained LLM weights stay frozen in every step, and
each step updates only the modules marked as trainable. The Stage~I objectives
are defined in Appendix~\ref{app:formation-objectives}, and the Stage~II
objectives in Appendix~\ref{app:access-objectives}.

\input{figures/training-pipeline/figure_training_pipeline}

\paragraph{Optimization.}
Table~\ref{tab:optimization-settings} lists the optimizer settings. The
ratio-clipping width $\epsilon=0.2$ and the dual-clip constant $c=3$ follow
\citet{ye2020mastering}, and $\delta=10^{-6}$ only stabilizes the
normalization in Equation~\ref{eq:access-advantages}. The decoder
LoRA~\citep{hu2021lora} is attached to the query, key, value, output, and MLP
projections of the top four decoder layers, with rank $r=16$ and
$\alpha=32$. We set the learning rates, the LoRA configuration, the auxiliary
loss weights, and the gradient-norm limit based on our experience. The Global Weavers use a separate learning rate for the output
projection that maps their latents to the decoder embedding space. Each
Stage~I step makes one pass over its supervision. The random seed is
20260911.

\begin{table}[!htbp]
\centering
\small
\setlength{\tabcolsep}{5pt}
\caption{\textbf{Optimization settings for memory formation and access
learning.}}
\label{tab:optimization-settings}
\begin{tabular}{lll}
\toprule
Stage & Parameter group or setting & Value \\
\midrule
I: Global formation & Image and Relation Weavers & $10^{-6}$ \\
 & Output projection & $10^{-5}$ \\
I: Local formation & Source Selector, Twig, and Local Weaver & $10^{-6}$ \\
I: interface warm-up & Marker and request rows & $10^{-4}$ \\
II: access learning & Decoder LoRA & $10^{-6}$ \\
 & Decoder LoRA rank $r$ / $\alpha$ & $16$ / $32$ \\
 & Decoder LoRA layers & top 4 of 28 \\
 & Request rows & $3\times10^{-4}$ \\
 & Ratio-clipping width $\epsilon$ & $0.2$ \\
 & Dual-clip constant $c$ & $3$ \\
 & Answer-loss weight $\lambda_{\mathrm{CE}}$ & $0.1$ \\
 & Preference-loss weight $\lambda_{\mathrm{pref}}$ & $0.1$ \\
All stages & Optimizer / weight decay & AdamW / $0$ \\
 & Maximum gradient norm & $1$ \\
\bottomrule
\end{tabular}
\end{table}

\paragraph{Rollouts and training length.}
Stage~II runs on 16 A800 GPUs with one prompt per GPU and no gradient
accumulation, so each step schedules 16 prompts. Each prompt samples four
root trajectories at temperature 1, and each eligible root adds two siblings,
so a step samples 64--192 natural trajectories. Continuation controls and
preference probes are sampled in addition. Rollouts use a 768-token
ordinary-text budget, a 64-token answer-completion grace period, and at most
five memory requests. These rollout sizes and budgets are set empirically.
The 40,000-example schedule contains 2,500 steps.

\paragraph{Checkpoint selection.}
We evaluate the model every 100 steps and at the end of training on a fixed
development set whose task list is set before training. The selected
checkpoint maximizes the equally weighted mean of per-task rewards, and ties
select the earlier checkpoint. The same weights are used for all final
benchmarks, and final benchmark scores do not enter the selection.

\subsection{Training Data}
\label{app:training-data}

\paragraph{Sources and allocation.}
We construct 50{,}000 image--question examples for Local memory formation,
30{,}000 for Global memory formation, and 40{,}000 for Stage~II access
learning. Each example contains one or more images, a question, and a
reference answer. Table~\ref{tab:app-training-sources} lists the source
datasets. All examples come from the training splits of these sources and are
deduplicated against every evaluation unit in
Table~\ref{tab:app-evaluation-splits}, so there is no leakage between training
and evaluation data.

\begin{table}[!htbp]
\centering
\small
\setlength{\tabcolsep}{5pt}
\renewcommand{\arraystretch}{1.12}
\caption{\textbf{Training data sources.} Source datasets provide
question--answer pairs, region annotations, or instance annotations from
which additional questions are constructed.}
\label{tab:app-training-sources}
\begin{tabular}{>{\raggedright\arraybackslash}p{0.32\linewidth}>{\raggedright\arraybackslash}p{0.60\linewidth}}
\toprule
Source & Content and use \\
\midrule
Visual-CoT (GQA)~\citep{shao2024visualcot}
& Object, attribute, and spatial-relation questions with region annotations. \\
COCO train2017~\citep{lin2014microsoft}
& Multi-image questions built from instance annotations. \\
Mantis-Instruct~\citep{jiang2024mantis}: ChartQA, Spot-the-Diff,
ImageCoDe, NLVR2, and Birds-to-Words
& Chart reading and arithmetic, image comparison, description-to-image
matching, and cross-image relational judgments. \\
Q-Zoom's VStar-COCO data~\citep{shi2026qzoom}
& Spatial-relation questions on COCO train2017 images. \\
ScanQA via M4-Instruct~\citep{li2024llavanextinterleave}
& Multi-view scene question answering, restricted to examples from the
original ScanQA training split. \\
\bottomrule
\end{tabular}
\end{table}

\paragraph{Stage~I annotations.}
Each Stage~I example carries a reference CoT, per-image facts, cross-image
relations, the answer, and target regions. Qwen3-VL-32B-Instruct generates
the reference CoTs. Answers come from the source datasets. Target regions
come from source region annotations and from model-generated step
annotations, which are checked against the source images and the proposed
crops. Per-image facts and cross-image relations are derived from source
annotations and from model-generated pseudo-labels. Facts and relations
supervise what Global memory encodes, region annotations supervise Local
source selection and localization, and answers supervise task completion.

\paragraph{Student-state targets.}
For student-state adaptation, we roll out Qwen2.5-VL-7B-Instruct and record
each text-and-latent prefix with its source selections, crops, and memory
returns. The teacher reviews each prefix together with its crop and returns
only the ROI target $\bar A$ and the continuation target $Y$. We do not reuse
the full teacher CoT as the continuation of a student state, because that CoT
need not follow from the student's prefix.

\paragraph{Access-learning data.}
Stage~II examples cover single-image perception, multi-image reasoning, and
general visual question answering. Rewards and the post-call answer loss use
the source answers, and the Stage~I memory annotations are not used in
Stage~II.

\subsection{Evaluation Splits and Scoring}
\label{app:eval-splits}

The evaluation suite contains ten benchmark families, divided into fifteen
evaluation units with 19{,}913 examples in total
(Table~\ref{tab:app-evaluation-splits}). Paired comparisons evaluate both
conditions on the same examples in every unit. BLINK uses its labelled
validation split, and MathVista uses test-mini. Mechanistic subsets and
development probes are reported separately from these evaluation units.

\begin{table}[!htbp]
\centering
\small
\setlength{\tabcolsep}{5pt}
\caption{\textbf{Evaluation units and aggregation rules.} Scores are reported
as percentages. The example count is the full evaluation denominator.}
\label{tab:app-evaluation-splits}
\begin{tabular}{lrl}
\toprule
Evaluation unit & Examples & Aggregated metric \\
\midrule
\multicolumn{3}{l}{\textit{Single-image perception}} \\
V*Bench, released test       &   191 & Micro accuracy \\
CV-Bench, test-2D            & 1{,}438 & Mean source accuracy \\
CV-Bench, test-3D            & 1{,}200 & Micro accuracy \\
HR-Bench, 4K                 &   800 & Mean category accuracy \\
HR-Bench, 8K                 &   800 & Mean category accuracy \\
\midrule
\multicolumn{3}{l}{\textit{Multi-image reasoning}} \\
BLINK, labelled validation   & 1{,}901 & Mean subtask accuracy \\
MuirBench, released test     & 2{,}600 & Micro accuracy \\
MMSI-Bench, non-circular set & 1{,}000 & Micro accuracy \\
MIMIC, Common test           & 1{,}000 & Accuracy \\
MIMIC, Counting test         & 5{,}000 & Accuracy \\
MIMIC, Odd-One-Out test      & 1{,}000 & Accuracy \\
MIMIC, Listing test          & 1{,}000 & Corpus micro set F1 \\
\midrule
\multicolumn{3}{l}{\textit{Broader visual understanding}} \\
MM-Vet v1, released set      &   218 & Mean fractional judge credit \\
MathVista, test-mini         & 1{,}000 & Accuracy \\
RealWorldQA, released test   &   765 & Micro accuracy \\
\midrule
Total                        & 19{,}913 & \\
\bottomrule
\end{tabular}
\end{table}

\paragraph{Aggregation.}
CV-Bench-2D averages the accuracies of its two sources with equal weight.
Each HR-Bench split averages four category accuracies, and BLINK averages its
fourteen subtask accuracies. These macro averages differ from micro accuracy
when components differ in size. MIMIC keeps its released metrics: accuracy for
Common, Counting, and Odd-One-Out, and corpus-level micro set F1 for Listing,
which is not an average of per-example F1 scores. The MIMIC Avg.\ column in
the main tables is the arithmetic mean of these four scores and is not an
official MIMIC metric. Unresolved predictions stay in the denominator and
receive no credit.

\paragraph{Judged benchmarks.}
MM-Vet is scored by the mean fractional credit from a fixed DeepSeek-V4-Flash
judge with a fixed JSON prompt. MathVista uses deterministic scoring for
multiple-choice answers and the same judge for mathematical equivalence of
free-form answers. These judges replace the online GPT-based judging of the
original benchmarks.

\paragraph{Comparison protocol.}
All methods in Table~\ref{tab:main-results} are evaluated on the same
examples with the same released questions, answer options, image order,
answer-format instructions, scoring rules, and judge. Each baseline uses its
authors' released checkpoint, tools, stopping rules, and generation caps and is
not retrained on our data; single-image methods receive an ordered contact
sheet for multi-image inputs. Because baselines differ in backbone and
training data, the main table also reports the backbone controls Native and
Coarse and the same-data controls SFT and Stage~I$_{\mathrm{sft}}$, which share
ReMAP's backbone and training data.

\paragraph{Prompts and generation settings.}
Native-protocol comparisons keep the released question, answer options, image
order, and answer-format instruction. Baselines that accept only one image
receive an ordered contact sheet of all input images, marked with a dagger in
Table~\ref{tab:app-mimic}. Generation budgets, stopping rules, and the
memory-request cap are reported with each evaluation condition.

\paragraph{Additional baseline configurations.}
MINT-CoT and ICoT use their Qwen2-VL-7B implementations, and Pixel Reasoner
uses its released Qwen2.5-VL-7B policy. ICoT keeps all input images but
replays visual tokens from the last image through its released selection
rule. Pixel Reasoner uses a 65{,}536-token runtime context for multi-image
inputs, caps each crop at 1{,}024 visual tokens and each generation round at
1{,}024 output tokens, keeps a 5{,}120-token cap on initial images, and
preserves its released crop tools and stopping rules. Q-Zoom caps its initial
image or ordered contact sheet and each RoI at 576 visual tokens, with at most
576 output tokens. These caps apply to each input or crop, not to the whole
trajectory.

\paragraph{Cross-backbone evaluation.}
\label{app:cross-backbone-protocol}
In Table~\ref{tab:cross-backbone}, Base denotes the original
backbone, OFF the trained model with Global and Local requests disabled, and
ReMAP the full model. The six benchmark columns are computed as $100n/N$ from
the correct counts $n$, with $N=191,1200,800,1901,2600,1000$ in column order.
For HR-Bench-8K and BLINK, these are micro accuracies rather than the
official four-category and fourteen-subtask macro averages used in the main
table. For every backbone, MIMIC Avg.\ is the mean over the four MIMIC
subsets. Differences are computed from unrounded ratios and rounded to two
decimals.

\paragraph{Paired scoring.}
The mechanistic comparisons share source examples, decoding settings, and
scoring rules; each study specifies its memory interface, request positions,
and content controls. Every expected example stays in the denominator, and an
unresolved prediction is scored as wrong. We report aggregate accuracy and
paired \textsc{help}/\textsc{harm} counts. Confidence intervals use a
sample-clustered bootstrap that resamples all states of an example together.
The full-cohort visual-influence and static pixel-budget analyses use 5{,}000
percentile resamples with seed 20260828, and the state-conditioned utility and
content-control analyses use 4{,}000 resamples with seed 20260830.
Budget-sweep intervals are pointwise across the five tested values of $K$.

\paragraph{MIMIC components.}
Table~\ref{tab:app-mimic} lists the four MIMIC component scores behind the
MIMIC Avg.\ column of the main table.

\begin{table}[!htbp]
\centering
\caption{\textbf{MIMIC component scores.} A dagger denotes an ordered
contact-sheet adaptation. VisMem is the public-code reproduction used in the
main table.}
\label{tab:app-mimic}
\begin{tabular}{lccccc}
\toprule
Method & Common & Counting & Odd-One-Out & Listing & Mean \\
\midrule
Native       & 62.60 & 44.98 & 60.40 & 12.69 & 45.17 \\
Coarse       & 61.10 & 38.42 & 54.70 & 20.02 & 43.56 \\
\midrule
Q-Zoom$^{\dagger}$ & 51.90 & 37.44 & 54.80 & 25.00 & 42.29 \\
DeepEyes$^{\dagger}$  & 52.60 & 40.38 & 53.20 & 10.90 & 39.27 \\
Pixel Reasoner & 64.50 & 43.34 & 30.00 & 41.46 & 44.83 \\
ReGround     & 53.60 & 36.76 & 13.00 &  8.66 & 28.01 \\
ICoT         & 48.80 & 15.38 & 54.90 &  0.55 & 29.91 \\
MINT-CoT     & 54.30 & 30.06 & 18.50 & 33.33 & 34.05 \\
DMLR$^{\dagger}$      & 38.70 & 16.14 & 17.70 & 22.41 & 23.74 \\
VaLR         & 38.10 & 12.38 & 20.60 &  0.62 & 17.93 \\
V-Reflection & 85.10 & 31.04 & 58.20 & 38.28 & 53.16 \\
VisMem       & 60.30 & 37.82 & 55.10 & 11.55 & 41.19 \\
\midrule
ReMAP        & 92.00 & 71.76 & 80.00 & 28.24 & 68.00 \\
\bottomrule
\end{tabular}
\end{table}

\subsection{Visual Influence and State-Conditioned Utility}
\label{app:bottleneck}

The two probes in this section measure how much the initial images still
affect predictions as reasoning proceeds, and whether the value of renewed
visual access depends on the reasoning state. Both use a fixed
Q-Zoom-adapted Qwen2.5-VL-7B checkpoint.

\subsubsection{Visual Influence Decay}

For each example, we generate one deterministic reasoning trajectory. We then
teacher-force the same text while replacing all input images with a
same-layout patch-shuffled or blank version, and compare the next-token
distributions $p_t$ and $\widetilde p_t$ under the original and perturbed
images:
\begin{equation}
  f_t=\mathbf{1}\!\left[\arg\max p_t\neq\arg\max\widetilde p_t\right],
  \qquad
  d_t=\operatorname{JS}(p_t,\widetilde p_t).
  \label{eq:visual-influence}
\end{equation}
The flip indicator $f_t$ records a change of the most likely next token, and
the Jensen--Shannon divergence $d_t$ also records changes that keep that
token. Positions are normalized by trajectory length and pooled into eight
bins. Both statistics measure the size of the visual influence, not its
direction. Figure~\ref{fig:app-visual-influence} plots the flip rate across
the eight bins, and Table~\ref{tab:app-influence} compares the first and last
parts of each trajectory.

\input{figures/visual-influence/figure_visual_influence}

\begin{table}[!htbp]
\centering
\caption{\textbf{Patch-shuffle visual influence across reasoning.}
Eligible/selected counts the trajectories analyzed out of the sampled
examples. First $\to$ last bin gives the flip rate in the first and eighth
bins. Paired change is the per-example mean flip rate over the last quarter
of the trajectory minus that over the first quarter, with a 95\% bootstrap
interval.}
\label{tab:app-influence}
\begin{tabular}{lccc}
\toprule
Cohort & Eligible / selected & First $\to$ last bin & Paired change \\
\midrule
V*Bench   & $189/191$ & $15.43\%\to2.36\%$ & $-8.79$ [$-10.48,-7.00$] pp \\
BLINK     & $254/256$ & $13.27\%\to2.45\%$ & $-10.88$ [$-12.15,-9.63$] pp \\
MuirBench & $254/256$ & $ 7.43\%\to1.96\%$ & $-4.50$ [$-5.48,-3.55$] pp \\
\midrule
Pooled    & $697/703$ & $11.73\%\to2.25\%$ & $-7.98$ [$-8.78,-7.21$] pp \\
\bottomrule
\end{tabular}
\end{table}

Blank images give the same pattern, with the flip rate falling from 14.82\%
to 2.50\%. Under patch shuffling, the mean JS divergence falls from 0.0320 to
0.0039. The signed change in the log probability of the generated token is
not consistent across cohorts: pooled, it is $-0.0211$ with interval
$[-0.0450,0.0026]$, and V*Bench has the opposite sign from BLINK and
MuirBench. The initial images therefore influence later predictions less,
but removing them does not uniformly lower the probability of the generated
text.

\subsubsection{State-Conditioned Utility}

We insert visual content at selected states of the same deterministic
trajectory and compare the gold-answer log probability. Candidate states are
the prompt-only state and the positions at 20\%, 40\%, 60\%, and 80\% of the
chain of thought. All arms share the textual prefix, the native visual
user-turn interface, and a 16-token return budget. The pair of states in each
contrast is chosen on development examples, and all estimates and intervals
in Table~\ref{tab:app-state-utility} are computed on an independent
confirmation split.

\begin{table}[!htbp]
\centering
\caption{\textbf{State and content contrasts in gold-answer log probability.}
Late--early contrasts insert the same content at the later and the earlier
selected state. Content contrasts compare the correct region or scene with a
random wrong region, a matched wrong scene, or a shuffled scene. NS marks an
interval that contains zero.}
\label{tab:app-state-utility}
\resizebox{\linewidth}{!}{%
\begin{tabular}{llcc}
\toprule
Cohort & Contrast & Estimate & 95\% CI \\
\midrule
V*Bench ($n=125$) & late--early correct-content utility
  & $+0.1771$ & $[+0.1327,+0.2227]$ \\
V*Bench ($n=125$) & correct--random-wrong RoI
  & $-0.0112$ & $[-0.0754,+0.0486]$ NS \\
CV-Bench-3D ($n=152$) & late--early current-scene utility
  & $+0.0726$ & $[+0.0405,+0.1050]$ \\
CV-Bench-3D ($n=152$) & current--matched-wrong scene
  & $+0.2265$ & $[+0.1361,+0.3171]$ \\
CV-Bench-3D ($n=152$) & current--shuffled scene
  & $+0.0264$ & $[-0.0358,+0.0882]$ NS \\
\bottomrule
\end{tabular}%
}
\end{table}

The same correct content is more useful at the later state on both V*Bench
($+0.1771$) and CV-Bench-3D ($+0.0726$), so the value of identical evidence
depends on when it is accessed. On CV-Bench-3D, the current scene is also
more useful than a matched wrong scene ($+0.2265$). The contrasts against a
random wrong region on V*Bench and against a shuffled scene on CV-Bench-3D
are not significant.

\subsection{Memory-Content Requirements and Controls}
\label{app:content-controls}

The two probes in this section test which memory content helps: freshly
encoded pixels of a region versus features from the initial encoding, and
correct relational context across images versus content-matched controls.
Both use the fixed checkpoint of Appendix~\ref{app:bottleneck}.

\subsubsection{Same-Box Pixel Re-encoding}

On the 191 V*Bench examples, both arms use the gold RoI, the same insertion
point in the prefill, and the same number $K$ of inserted tokens. The feature
arm pools the RoI features from the initial encoding. The pixel arm crops the
region from the source image, encodes it at high resolution, and compresses
the result to $K$ tokens. Figure~\ref{fig:app-pixel-reencoding} reports five
values of $K$.

\input{figures/content-controls/figure_pixel_reencoding}

Pixel re-encoding matches the feature arm at $K=4$ ($-0.0007$ in gold-answer
log probability) and exceeds it from $K=8$ onward. At $K=32$, it raises the
log probability by $0.2243$ (95\% CI $[0.1241,0.3325]$) and accuracy from
62.30\% to 71.73\%. The
high-resolution RoI pass uses 278.4 visual tokens on average, compared with
1{,}244.4 for the high-resolution full image. At $K=16$, pixel memory also
exceeds wrong-RoI, shuffled, blank, and zero controls. Splitting the examples
into target-size tertiles gives the same direction in every tertile, with
small differences between tertiles.

\subsubsection{Relational-Role Probe}

We build 192 non-overlapping part--whole image pairs from COCO train2017
instance annotations, covering 800 unique images. In the role-changing
condition, the focal image stays pixel-identical while its relation to the
other image changes. In the appearance control, the appearance changes while
the focal role stays fixed. Sensitivity is the total-variation distance
between the model's answer distributions before and after a change, and
fixation is the rate at which the model keeps predicting the focal option
after the roles are swapped.

\begin{table}[!htbp]
\centering
\small
\caption{\textbf{Relational-role probe.} The upper rows use total-variation
sensitivity or fixation rate on 192 pairs. The lower rows compare the oracle
relational context with each control in gold-answer log probability on the
71 selected late failure states. NS marks an interval that contains zero.}
\label{tab:app-relation}
\begin{tabular}{lcc}
\toprule
Contrast & Estimate & 95\% CI \\
\midrule
Role change $-$ appearance control
  & $+0.0482$ & $[+0.0254,+0.0725]$ \\
Late $-$ early role sensitivity
  & $+0.0039$ & $[-0.0056,+0.0138]$ NS \\
Late $-$ early fixation rate
  & $+0.0182$ & $[-0.0026,+0.0391]$ NS \\
\midrule
Oracle $-$ no context        & $+2.5797$ & $[+2.1222,+2.9881]$ \\
Oracle $-$ blank block       & $+2.4765$ & $[+2.1251,+2.8061]$ \\
Oracle $-$ zero block        & $+2.8028$ & $[+2.5116,+3.0891]$ \\
Oracle $-$ object list only  & $+1.3852$ & $[+1.1268,+1.6433]$ \\
Oracle $-$ another sample    & $+1.2927$ & $[+1.0276,+1.5654]$ \\
Oracle $-$ wrong role        & $+3.1493$ & $[+2.7849,+3.5274]$ \\
\bottomrule
\end{tabular}
\end{table}

The model is more sensitive to a role change than to an appearance change
($+0.0482$), so its predictions depend on cross-image relations and not only
on the focal image (Figure~\ref{fig:app-relational-role}(a)). Neither role
sensitivity nor fixation changes significantly between early and late states.

\input{figures/content-controls/figure_relational_role}

To test whether correct relational context can repair these errors, we
select the 71 late states at which the model predicts the focal option for
both the true and the role-swapped prefix, so that its role-swapped
prediction is wrong. At each state, we insert either the oracle relational
context or one of six controls: no context, a blank block, a zero block, an
object list only, another sample's context, or the wrong role. Without
context, all 71 states are wrong by construction. The oracle context repairs
41 of them, an accuracy of 57.7\%, compared with 53.5\%, 52.1\%, 47.9\%, 49.3\%, and 35.2\% for the blank, zero,
object-list, other-sample, and wrong-role controls
(Figure~\ref{fig:app-relational-role}(b)). In gold-answer log probability, it exceeds every control
(Table~\ref{tab:app-relation}). Because the states are selected as failures
and the relation comes from an oracle, this probe measures what correct
relational content can contribute, not the effect of the learned Global
memory.

\subsection{Additional Ablations}
\label{app:additional-ablations}
\input{sections/90_ablation_protocol}

\subsection{Branch-Point Selection Diagnostic}
\label{app:branch-selection-diagnostic}

This diagnostic compares four rules for choosing the state at which a root
trajectory branches, with the policy frozen. The rules are the score
$u(s_t,a_t)$ in Equation~\ref{eq:crd}, labeled CRD in
Figure~\ref{fig:branch-selector-ablation}; the route entropy
$H\big(p_t(\mathrm N),p_t(\mathrm G),p_t(\mathrm L)\big)$; uniform random
selection among eligible states; and the entropy-change signal of
ARPO~\citep{dong2025arpo}. The ARPO-style signal is the mean entropy of the
last four ordinary tokens minus the mean entropy of the first four generated
tokens. We branch at its maximum under the shared budget instead of using
ARPO's stochastic branching gate. Unlike the route entropy, $u(s_t,a_t)$
conditions on the realized route and measures how likely a sibling is to
depart from it.

All rules receive the same root trajectories and eligible candidate states.
The cohort contains 192 training prompts, 32 from each of six ability
categories: attribute, counting, existence, local detail, relation, and
spatial reasoning. The prompts are selected without using rewards or
correctness. Each prompt has four root trajectories, and each root yields a
family of the original continuation and two siblings resampled from the
selected state, giving 12 trajectories per rule and prompt.

Routing Yield@H16 is the fraction of families whose trajectories produce
different sets of Global and Local requests within the first 16 generated
positions after the branch. Joint EAY multiplies this diversity indicator by
the mean absolute deviation of the family's final-answer rewards, so it is
positive only when routing diversity comes with different rewards. Both
metrics are averaged over the four families of a prompt, and we report
prompt means. Confidence intervals use 10{,}000 prompt-level bootstrap
resamples stratified by ability category, with seed 26090601 and the same
resamples for all rules.

The score $u(s_t,a_t)$ gives the highest routing yield, 75.91\% (95\% CI
$[72.92,78.78]$), compared with 68.36\% $[65.23,71.48]$ for the route
entropy, 38.54\% $[35.03,41.93]$ for random selection, and 32.16\%
$[28.65,35.68]$ for the ARPO-style signal. Its Joint EAY of 0.144
$[0.131,0.159]$ is close to that of the route entropy, 0.143
$[0.129,0.157]$, and above those of random selection, 0.084, and the
ARPO-style signal, 0.067. Route-based selection therefore exposes more
alternative memory decisions at the branch, and these alternatives come with
larger reward differences than those found by random or entropy-change
selection.

\input{figures/branch-ablation/figure_branch_ablation}

\subsection{Complete Qualitative Case Traces}
\label{app:complete-case-traces}

Figure~\ref{fig:visual-evidence-acquisition-appendix} shows the complete
reasoning of the two main-text cases, with and without memory.
\input{figures/experiment-analysis/visual_evidence_acquisition_appendix}

\subsection{Visual-Context Accounting}
\label{app:visual-context-protocol}

Figure~\ref{fig:memory-invocation}(a) compares the visual tokens that enter
the decoder sequence with the performance in
Table~\ref{tab:main-results}. For each question, we count the initial
image tokens and every crop token or memory latent returned during reasoning.
A repeated Global request inserts the cached block again, and its latents are
counted again. Ordinary text tokens, request and end markers, and
representations used only inside the vision encoder or the memory weavers are
excluded. The count therefore measures the visual context of the decoder, not
total inference compute.

The comparison uses the questions measured for all five methods: 78 V*Bench,
60 CV-Bench-2D, and 107 MuirBench questions. Means are unweighted over these
questions and include questions without memory requests. The performance
lines use the full-benchmark scores of the main table. ReMAP starts from at
most 1{,}024 visual tokens per image and returns 32 latents per Local request
and $4N+4$ latents per Global request. Its counts come from the recorded
memory insertions of its evaluation run. For Q-Zoom and DeepEyes on
MuirBench, the counts sum the passes over each input image.

Table~\ref{tab:app-visual-context} lists the per-question means. ReMAP adds
16.0, 18.7, and 29.3 tokens per question through memory returns, and it uses
76.8\%, 51.0\%, and 52.3\% fewer visual tokens than Native on V*Bench,
CV-Bench-2D, and MuirBench, respectively.

\input{tables/visual_context_accounting}

%% file: sections/92_related_work.tex
\subsection{Related Work and Positioning}
\label{app:related-work}

\paragraph{Memory curation: acquiring and selecting visual evidence.}
Memory curation determines which evidence reasoning acquires. DeepEyes learns
crop requests \citep{zheng2026deepeyes}, while Pixel Reasoner combines image
cropping and video-frame selection \citep{su2025pixelreasoner}. VLM-R$^3$
interleaves region selection, cropping, and zooming with textual reasoning
\citep{jiang2025vlmr3}. VPT supports
token-guided cropping and state-conditioned full-image re-encoding
\citep{yu2025vpt}. CMMCoT generates image indices and boxes to re-encode entity
regions during multi-image reasoning \citep{zhang2026cmmcot}. Qwen Look Again,
ICoT, and MINT-CoT instead select or reuse encoded visual tokens
\citep{chu2025lookagain,gao2025interleaved,chen2025mintcot}. VTPerception-R1
grounds reasoning in explicit perceptual descriptions \citep{ding2025vtperception},
while perception-time scaling decomposes perceptual estimates into symbolic
intermediate steps \citep{li2026perceptiontimescaling}. VisMem uses
state-conditioned formers to produce perceptual and semantic memory
\citep{yu2026vismem}. V-Reflection trains latent reasoning states with spatial
supervision \citep{zhou2026vreflection}.

\paragraph{Memory organization: representing visual evidence.}
Visual Sketchpad represents intermediate evidence as generated sketches
\citep{hu2024visualsketchpad}, VTool-R1 as annotated images
\citep{wu2025vtoolr1}, and CSMR as text \citep{zhang2026lookondemand}.
Take-along Visual Conditioning pools visual tokens \citep{sun2025takealong},
while Persistent Visual Memory projects image embeddings into a lower-dimensional
space \citep{huang2026persistent}. CMMCoT retrieves cross-image cues from cached
per-layer visual key--value features \citep{zhang2026cmmcot}. FlexMem compresses and retrieves visual KV
memories for long-video understanding \citep{chen2026flexmem}. MemGen and VisMem
represent memory as latent token sequences
\citep{zhang2026memgen,yu2026vismem}. Laser and Mirage represent intermediate
reasoning through continuous hidden states or self-generated visual tokens
\citep{wang2026forest,yang2026machine}. MCOUT-Multi refines continuous thoughts by
attending to multimodal input embeddings \citep{pham2025mcout}, while Cognitive
Supersensing predicts answer-image latents and uses them to guide rationale
training \citep{li2026cognitive}. IVT-LR combines recurrent hidden states with
selected input-image embeddings for latent reasoning \citep{chen2026reasoningdark}.
ILVR interleaves text with generated visual latents supervised by selected
features from intermediate helper images \citep{dong2026ilvr}. UniVLR supervises
latent reasoning with visual encodings of rendered reasoning text and auxiliary
images \citep{jiang2026univlr}. \citet{zhang2026unsilencing} identify underused
visual latents and optimize them at inference time with a frozen backbone.

\paragraph{Memory access: timing and selecting memory use.}
Take-along Visual Conditioning and Qwen Look Again reactivate visual evidence
at reflection events \citep{sun2025takealong,chu2025lookagain}. ReGround uses
self-diagnosis to guide re-examination \citep{peng2026reground}, and VGR triggers
region-specific replay from precomputed visual features \citep{wang2026vgr}.
DaP-ICoT inserts selected object
images when the reasoning step's mean logit margin falls below a threshold
\citep{liu2026dapicot}. MemGen learns a memory trigger \citep{zhang2026memgen}.
LatentMorph learns when to invoke latent reasoning over compact visual memory
during image generation \citep{chen2026latentmorph}.
VisMem inserts perceptual or semantic latent memory through invocation
tokens \citep{yu2026vismem}. Its two memories differ in abstraction level
rather than in spatial scope, and a request is not compared with continuing
from the same state. CSMR invokes a separate perception module
\citep{zhang2026lookondemand}. For policy training, ARPO allocates
entropy-guided rollout branches after tool feedback \citep{dong2025arpo}.
PEPO combines visual hidden-state similarity and token entropy to reweight
token advantages without auxiliary branches \citep{li2026rethinking}.

\paragraph{Positioning ReMAP.}
ReMAP bridges these dimensions by connecting active visual evidence acquisition
with latent memory and learned access. Within memory curation, it emphasizes
the complementarity between newly acquired local details and the scene and
cross-image context needed to interpret them. Within memory organization, it
brings pixel-level reacquisition into compact latent memory, connecting renewed
perception with an efficient representation of visual evidence. Within memory
access, it focuses on the utility of requesting evidence at a given reasoning
state, using comparisons between invocation and continuation to guide selective
access. Together, these choices support sustained visual grounding throughout
reasoning.

\input{tables/related_work_positioning}

%% file: tables/related_work_positioning.tex
\definecolor{PositionGreen}{HTML}{228B22}
\definecolor{PositionRed}{HTML}{C00000}
\definecolor{PositionGray}{HTML}{777777}
\newcommand{\posyes}{\textcolor{PositionGreen}{\ding{51}}}
\newcommand{\posno}{\textcolor{PositionRed}{\ding{55}}}
\newcommand{\posna}{\textcolor{PositionGray}{--}}

\begin{table*}[t]
\centering
\small
\setlength{\tabcolsep}{2.5pt}
\renewcommand{\arraystretch}{1.12}
\caption{\textbf{Qualitative positioning of representative visual-memory methods.}
Green checks indicate explicit support, red crosses indicate non-core
mechanisms, and gray dashes indicate dimensions not emphasized in the cited
work.}
\label{tab:related-work-positioning}
\begin{tabular}{@{}>{\raggedright\arraybackslash}p{0.18\textwidth}*{6}{>{\centering\arraybackslash}p{0.12\textwidth}}@{}}
\toprule
Method & \parbox[c]{0.12\textwidth}{\centering\scriptsize Pixel / Region\\Reacquisition} & \parbox[c]{0.12\textwidth}{\centering\scriptsize Compact Visual\\Memory} & \parbox[c]{0.12\textwidth}{\centering\scriptsize State-\\Conditioned Access} & \parbox[c]{0.12\textwidth}{\centering\scriptsize Cross-Image\\Context} & \parbox[c]{0.12\textwidth}{\centering\scriptsize Global / Local\\Split} & \parbox[c]{0.12\textwidth}{\centering\scriptsize Invocation vs.\\Continuation} \\
\midrule
DeepEyes & \posyes & \posno & \posna & \posna & \posno & \posno \\
Pixel Reasoner & \posyes & \posno & \posna & \posna & \posno & \posno \\
VLM-R$^3$ & \posyes & \posno & \posna & \posna & \posno & \posno \\
CMMCoT & \posyes & \posyes & \posna & \posyes & \posno & \posno \\
Qwen Look Again & \posno & \posyes & \posna & \posna & \posno & \posno \\
ICoT / MINT-CoT & \posno & \posyes & \posna & \posna & \posno & \posno \\
MemGen & \posno & \posyes & \posyes & \posna & \posno & \posno \\
VisMem & \posno & \posyes & \posyes & \posna & \posno & \posno \\
FlexMem & \posno & \posyes & \posna & \posna & \posno & \posno \\
ReGround & \posyes & \posno & \posyes & \posna & \posno & \posno \\
\midrule
\rowcolor{ReMAPSecond}
\textbf{ReMAP (ours)} & \posyes & \posyes & \posyes & \posyes & \posyes & \posyes \\
\bottomrule
\end{tabular}
\end{table*}

%% file: figures/training-pipeline/figure_training_pipeline.tex
\begin{figure}[!htbp]
  \centering
  \includegraphics[width=\linewidth]{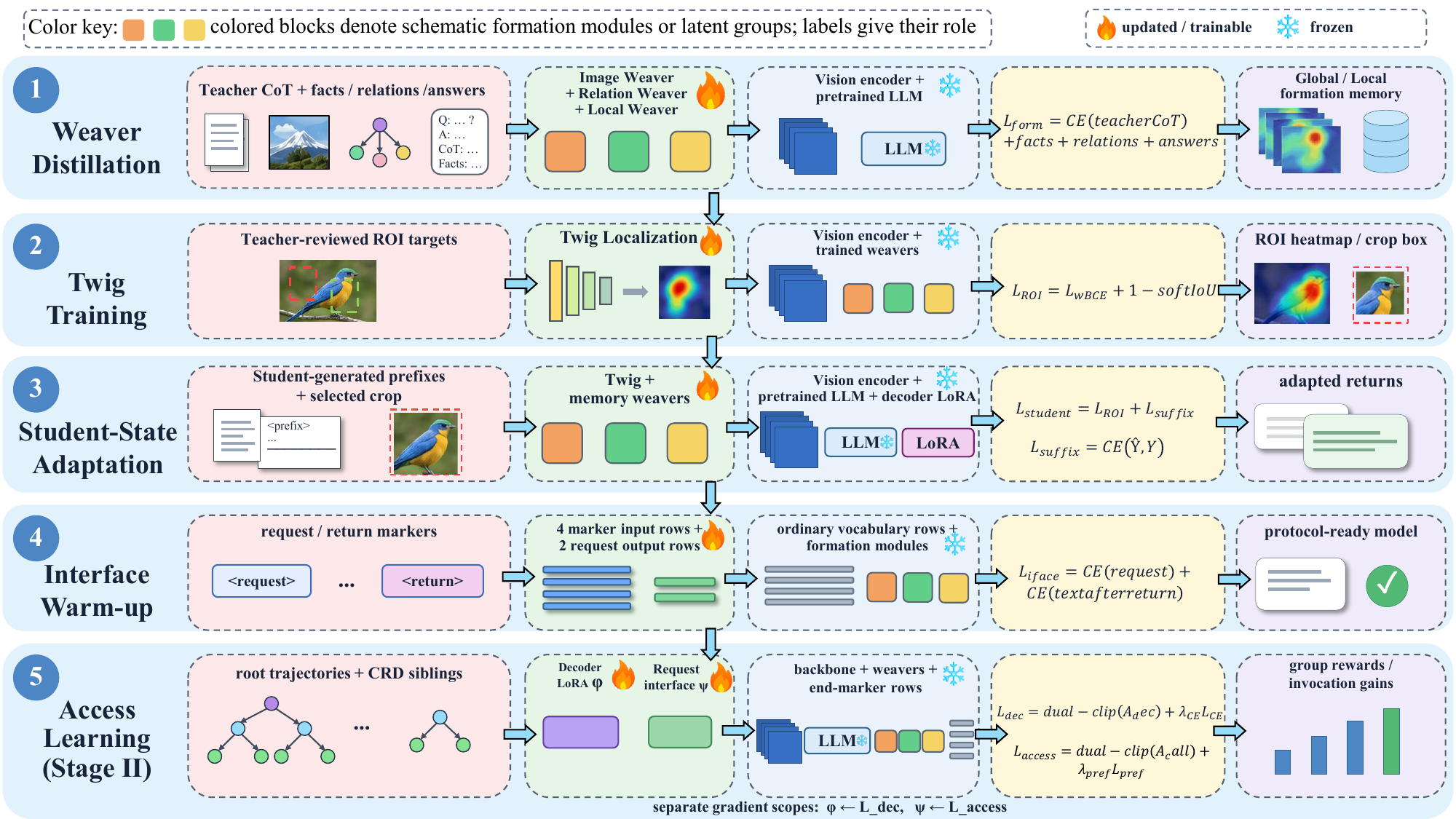}
  \caption{\textbf{Five-step training pipeline.} Each row lists the inputs of
  one step, the updated modules (flame), the frozen components (snowflake), the
  objective, and the output. Steps 1--4 form Stage~I, and step 5 is Stage~II
  access learning, in which the decoder LoRA $\phi$ and the request interface
  $\psi$ receive gradients only from $\mathcal L_{\mathrm{dec}}$ and
  $\mathcal L_{\mathrm{access}}$, respectively.}
  \label{fig:app-training-pipeline}
\end{figure}

%% file: figures/visual-influence/figure_visual_influence.tex
\begin{figure}[!htbp]
  \centering
  \includegraphics[width=\linewidth]{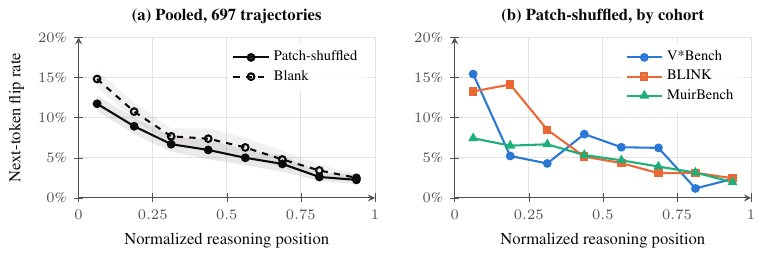}
  \caption{\textbf{Visual influence decays along the reasoning trajectory.}
  Rate at which the next-token top-1 decision changes when the input images
  are replaced, in eight bins of normalized reasoning position.
  (a) Patch-shuffled and blank images, pooled over V*Bench, BLINK, and
  MuirBench, with 95\% bootstrap intervals. (b) Patch-shuffled images for each
  cohort.}
  \label{fig:app-visual-influence}
\end{figure}

%% file: figures/content-controls/figure_pixel_reencoding.tex
\begin{figure}[!htbp]
  \centering
  \includegraphics[width=\linewidth]{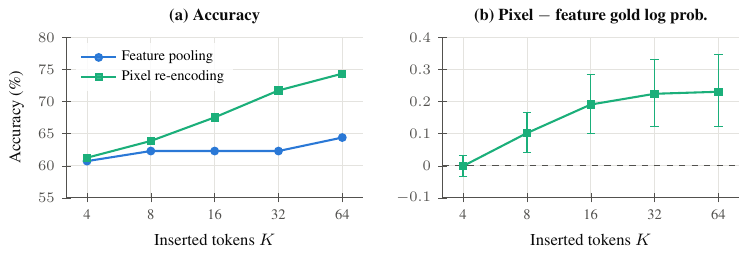}
  \caption{\textbf{Pixel re-encoding versus feature pooling on V*Bench.} Both
  arms return the gold RoI with $K$ inserted tokens at the same insertion point
  on 191 examples. (a) Accuracy. (b) Pixel minus feature gold-answer log
  probability, with 95\% CIs.}
  \label{fig:app-pixel-reencoding}
\end{figure}

%% file: figures/content-controls/figure_relational_role.tex
\begin{figure}[!htbp]
  \centering
  \includegraphics[width=\linewidth]{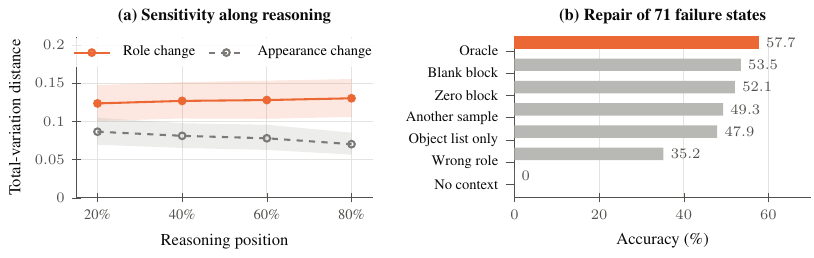}
  \caption{\textbf{Relational-role probe.} (a) Total-variation sensitivity to
  a role change and to an appearance change at four reasoning positions on 192
  pairs, with 95\% intervals. (b) Accuracy on the 71 selected late failure
  states with the oracle relational context and six controls; without context,
  every state is wrong by construction.}
  \label{fig:app-relational-role}
\end{figure}

%% file: sections/90_ablation_protocol.tex
\paragraph{Shared evaluation setting.}
The memory-curation and memory-access variants in
Table~\ref{tab:component-ablation} use the complete ReMAP weights
and share the input examples, preprocessing, scoring, and decoding limits.
Base is the Coarse backbone with at most 1{,}024 visual tokens per image. The
table uses all 191 V$^\star$Bench, 1{,}200 CV-Bench-3D, 2{,}600 MuirBench,
1{,}901 BLINK, and 218 MM-Vet examples. V$^\star$Bench, CV-Bench-3D,
MuirBench, and BLINK report micro accuracy, and MM-Vet reports the mean graded
score. BLINK in this table therefore uses micro accuracy rather than the
mean-subtask aggregation of Appendix~\ref{app:eval-splits}. Unresolved answers
count as incorrect.

\paragraph{Memory-curation variants.}
w/o Memory masks both request tokens. w/ Global memory and w/ Local memory
enable only one request type and keep the decoder adapter, request
parameters, and answer path of ReMAP. When Global requests are disabled, Local
still uses the cached Global memory $Z_G$, and the unused return capacity is
not reassigned. Each Local request returns 32 latent tokens, including on
multi-image examples. All variants use a 768-token ordinary-text budget, a
64-token answer grace period, and at most five memory requests. Memory
requests do not count toward the text budget.

\paragraph{Training variants.}
SFT distills the teacher CoT without memory, and Stage~I$_{\mathrm{sft}}$
distills both the CoT and the memory. ReMAP adds access learning after
Stage~I. These variants differ in training. w/o Memory instead shares all of
ReMAP's weights, including the Stage~II decoder LoRA, and masks Global and
Local requests only at inference. The gap between ReMAP and w/o Memory
therefore measures the contribution of reasoning-time memory access on top of
the full SFT and RL training with parameters held fixed; it is an
inference-time ablation, not a model trained without memory.

\paragraph{Invocation controls.}
Fixed, Random, and Entropy use the complete ReMAP weights and change only when
requests are issued at inference. Fixed requests Global at positions $0$ and
$\lfloor0.7B\rfloor$ and Local at $\lfloor0.3B\rfloor$ and
$\lfloor0.5B\rfloor$ of the text budget $B=768$, that is, at positions 0, 230,
384, and 537. Random and Entropy consider a request after each generated
delimiter, excluding position 0. Random issues a request with a fixed
probability, following the random insertion of MemGen and
VisMem~\citep{zhang2026memgen,yu2026vismem}; as in VisMem, it samples
requests at delimiters. Entropy issues a request when the entropy of the
next-token distribution over ordinary text tokens exceeds a fixed threshold.
Memory tokens are excluded from this distribution, and EOS is retained. A
triggered request chooses Global or Local with equal probability, as in
VisMem. In all three controls, committing an answer or emitting EOS cancels
later requests, and the five-request cap applies. Because each policy decides
its own requests, the number of requests differs across policies.

%% file: figures/branch-ablation/figure_branch_ablation.tex
\begin{figure*}[t]
  \centering
  \includegraphics[width=\textwidth]{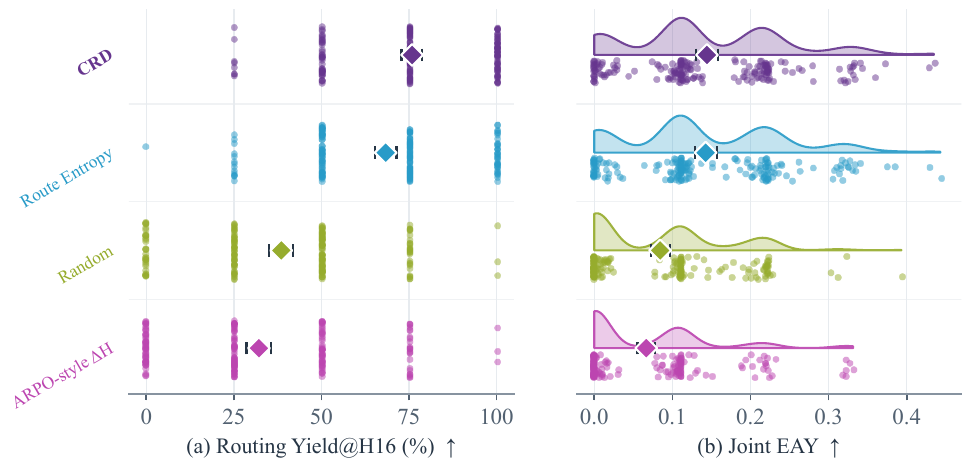}
  \caption{
    \textbf{Branch-point selection with a fixed trajectory budget.}
    (a) Routing Yield@H16, the fraction of branch families whose trajectories
    make different Global and Local requests within 16 positions after the
    branch. (b) Joint EAY, the reward variation that co-occurs with this
    routing diversity. Each point averages four root families of one prompt;
    diamonds and bars mark the mean and 95\% bootstrap interval. All rules use
    the same roots, candidate states, and 12 trajectories per prompt on 192
    prompts, with the policy frozen. ARPO-style $\Delta H$ is the
    fixed-budget adaptation of ARPO's entropy-change signal.
  }
  \label{fig:branch-selector-ablation}
\end{figure*}

%% file: figures/experiment-analysis/visual_evidence_acquisition_appendix.tex
\begin{figure}[p]
  \centering
  \includegraphics[width=\textwidth,height=0.86\textheight,keepaspectratio]{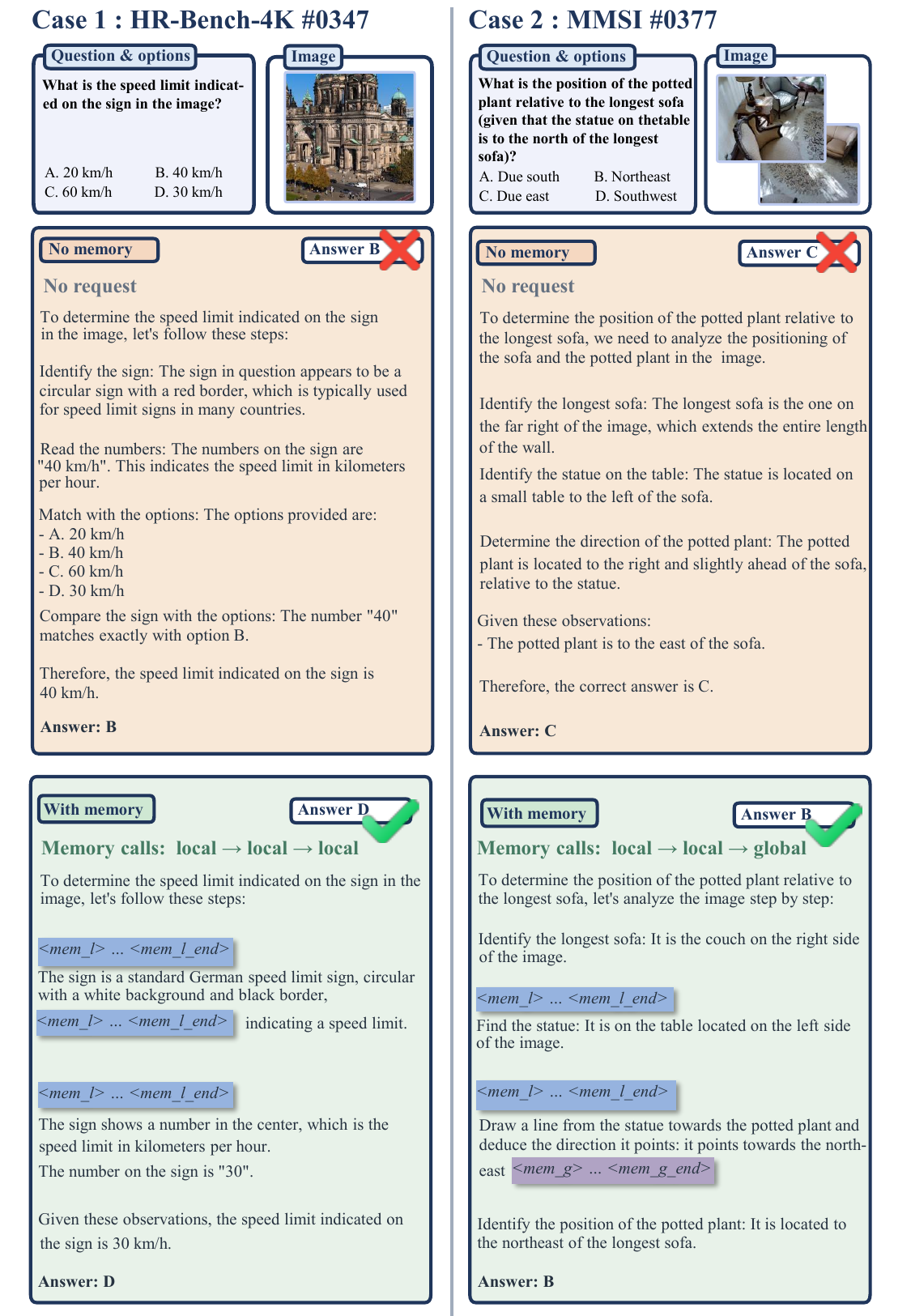}
  \caption{\textbf{Complete paired reasoning traces.} Cases~1 and~2 expand
  panels (a) and (b) of Figure~\ref{fig:visual-evidence-acquisition} with the
  full recorded reasoning of the paired no-call and memory-enabled
  trajectories. Each memory return is marked by its request and end tokens.
  Case~1 (HR-Bench-4K \#0347): without memory, the model reads 40~km/h and
  answers B; after three Local calls, it reads 30~km/h and answers D.
  Case~2 (MMSI \#0377): without memory, the model answers due east (C); after
  two Local calls and one Global call, it answers northeast (B).}
  \label{fig:visual-evidence-acquisition-appendix}
\end{figure}

%% file: tables/visual_context_accounting.tex
\begin{table}[!htbp]
\centering
\small
\setlength{\tabcolsep}{5pt}
\caption{\textbf{Visual tokens entering the decoder sequence per question.} Means over the questions shared by all five methods, with the question count given for each benchmark. Added counts crop tokens and memory latents returned during reasoning. Totals are computed before rounding.}
\label{tab:app-visual-context}
\begin{tabular}{llrrr}
\toprule
Benchmark & Method & Initial & Added & Total \\
\midrule
V*Bench ($n=78$) & Native & 4,417.9 & 0.0 & 4,417.9 \\
 & Q-Zoom & 1,008.0 & 331.8 & 1,339.8 \\
 & DeepEyes & 4,417.9 & 35.9 & 4,453.8 \\
 & VisMem & 4,417.9 & 0.0 & 4,417.9 \\
 & ReMAP & 1,008.0 & 16.0 & 1,024.0 \\
\midrule
CV-Bench-2D ($n=60$) & Native & 906.0 & 0.0 & 906.0 \\
 & Q-Zoom & 450.6 & 175.3 & 625.9 \\
 & DeepEyes & 906.0 & 208.1 & 1,114.1 \\
 & VisMem & 906.0 & 0.0 & 906.0 \\
 & ReMAP & 425.3 & 18.7 & 444.0 \\
\midrule
MuirBench ($n=107$) & Native & 7,099.5 & 0.0 & 7,099.5 \\
 & Q-Zoom & 3,515.4 & 500.8 & 4,016.2 \\
 & DeepEyes & 7,099.5 & 2,097.6 & 9,197.0 \\
 & VisMem & 7,099.5 & 0.0 & 7,099.5 \\
 & ReMAP & 3,359.9 & 29.3 & 3,389.2 \\
\bottomrule
\end{tabular}
\end{table}